\documentclass[journal]{IEEEtran}
\IEEEoverridecommandlockouts
\usepackage{autobreak}
\usepackage{amsmath,amsfonts}
\usepackage{bbm}
\usepackage[noend]{algpseudocode}
\usepackage[linesnumbered]{algorithm2e}

\usepackage{blindtext}
\usepackage{gensymb}
\usepackage{array}
\usepackage{textcomp}
\usepackage{stfloats}
\usepackage{url}
\usepackage{verbatim}
\usepackage{graphicx}
\usepackage[font=footnotesize]{caption}
\usepackage[caption=false,font=footnotesize]{subfig}
\usepackage{cite}
\usepackage{multirow}
\usepackage{color}
\usepackage{amsmath}

\definecolor{purple}{rgb}{0.44,0.01,0.55}
\definecolor{orange}{rgb}{0.79,0.375,0.08}
\definecolor{bluey}{rgb}{0,0.48,0.65}

\usepackage[compact]{titlesec}
\titlespacing{\section}{0pt}{0pt}{0pt}
\titlespacing{\subsection}{0pt}{*0}{*0}
\titlespacing{\subsubsection}{0pt}{*0}{*0}
\AtBeginDocument{
\setlength\abovedisplayskip{0pt}
\setlength\belowdisplayskip{0pt}}

\usepackage{xpatch}
\xpatchcmd{\IEEEbiography}
{\fontsize{8.25}{10}\selectfont}
{\normalsize}
{}{}

\begin{document}

\title{\Large{A Vision Toward Energy-Efficient Domain-Specific Artificial Intelligence Models and Agents}
\thanks{Abhijit Chatterjee, Niraj Jha, and Keshab K. Parhi contributed equally to this paper.}}
\author{
    \IEEEauthorblockN{
        Abhijit Chatterjee, Fellow, \textit{IEEE} \IEEEauthorrefmark{1};
        Niraj K. Jha, Fellow, \textit{IEEE} \IEEEauthorrefmark{2};
        Jonathan D. Cohen\IEEEauthorrefmark{3};
        Thomas L. Griffiths\IEEEauthorrefmark{4};
        Hongjing Lu\IEEEauthorrefmark{5};
        Diana Marculescu, Fellow, \textit{IEEE} \IEEEauthorrefmark{6};
        Ashiqur Rasul\IEEEauthorrefmark{1};
        Wenrui Xu, \IEEEauthorrefmark{7}; and 
        Keshab K. Parhi, Fellow, \textit{IEEE} \IEEEauthorrefmark{7}
    }
    
    \IEEEauthorblockA{\IEEEauthorrefmark{1}School of Electrical and Computer Engineering, Georgia Institute of Technology \\
    \texttt{Email: \{abhijit.chatterjee@ece., arasul6@\}gatech.edu}}
    
    \IEEEauthorblockA{\IEEEauthorrefmark{2}Department of Electrical and Computer Engineering, Princeton University \\
    \texttt{Email: jha@princeton.edu}}
    
    \IEEEauthorblockA{\IEEEauthorrefmark{3}Department of Psychology \& Neuroscience Institute, Princeton University \\
    \texttt{Email: jdc@princeton.edu}}
    
    \IEEEauthorblockA{\IEEEauthorrefmark{4}Departments of Psychology \& Computer Science, Princeton University \\
    \texttt{Email: tomg@princeton.edu}}
    
    \IEEEauthorblockA{\IEEEauthorrefmark{5}Department of Psychology, University of California, Los Angeles \\
    \texttt{Email: hongjing@ucla.edu}}
    
    \IEEEauthorblockA{\IEEEauthorrefmark{6}Department of Electrical and Computer Engineering, University of Texas at Austin \\
    \texttt{Email: dianam@utexas.edu}}
    
    \IEEEauthorblockA{\IEEEauthorrefmark{7}Department of Electrical and Computer Engineering, University of Minnesota, Twin Cities \\
    \texttt{Email: \{xu000424,parhi\}@umn.edu}}
}
\maketitle

\setcounter{page}{1}
\pagenumbering{arabic}

\begin{abstract}
The field of artificial intelligence (AI) has taken a tight hold on broad aspects of society, industry, business, and governance in ways that dictate the prosperity and might of the world's economies. The AI market size is projected to grow from {\$}189 billion in 2023 to {\$}4.8 trillion by 2033. Currently, AI is dominated by large language models (LLMs) that exhibit linguistic and visual intelligence. However, training these models requires a massive amount of data scraped from the web as well as large amounts of energy (50-60 GWh to train GPT-4). Despite these costs, these models often hallucinate, a characteristic that prevents them from being deployed in critical application domains. In contrast, the human brain consumes only 20W of power. What is needed is the next level of AI evolution in which lightweight domain-specific multimodal models, especially compact models with 10--20B parameters for bounded domains, with higher levels of intelligence can reason, plan, and make decisions in dynamic environments with real-time data and prior knowledge, while learning continuously and evolving in ways that enhance future decision-making capability. This will define the next wave of AI, progressing from today's large
models, trained with vast amounts of data, to nimble energy-efficient domain-specific agents that can reason and think in a world full of uncertainty. To support such agents, hardware will need to be reimagined to allow system-level energy efficiencies $\geq {1000X}$  over the state of the art for targeted domain tasks, subject to accuracy, latency, and coverage constraints. Such a vision of future AI systems is developed in this work.
\end{abstract}

\begin{IEEEkeywords}
AI models, AI agents, language models, learning, reasoning, energy-efficient AI, analogical reasoning, prospective learning, meta reasoning, sample efficiency, computational efficiency.
\end{IEEEkeywords}

\section{Introduction}

There has been a lot of excitement generated by recent single and multimodal large language models (LLMs), such as ChatGPT \cite{zhou2023comprehensive,kocon2023chatgpt}, DALL-E \cite{ramesh2022hierarchical,singh2021illiterate}, LaMDA \cite{thoppilan2022lamda}, PaLM-E \cite{driess2023palm}, GPT-4 \cite{bubeck2023sparks}, Llama \cite{touvron2023llama}, and Deepseek \cite{guo2025deepseek}. These models can answer sophisticated queries and perform text-to-image transformations in unprecedented ways, opening up a vast plethora of opportunities for industry. However, a key tenet of current machine learning algorithms is that the \textit{training and test data be drawn from the same probability distribution}. Therefore, in uncertain or unknown environments where correlations may not hold, the models are brittle and predictions are unreliable. Furthermore, the number of parameters in these models continues to grow significantly. As an example, 
GPT-3 has 175 billion parameters compared to 1.76 trillion parameters in GPT-4. However, despite this massive increase in model size, the underlying models struggle with brain-like cognition: the ability to \textit{learn continuously, reason, think, make decisions, and adapt} in a world full of uncertainty when faced with problems and situations not encountered before. 

The main claim of this paper is: \textit{domain-specific, energy-efficient AI models and agents are a more viable path for many critical applications than general-purpose LLMs}. In such settings, the objective is not to reproduce every capability of a frontier model. It is to achieve reliable reasoning, perception, and decision-making for a specific domain with much less training data, lower inference energy, and explicit links to trusted knowledge. A compact model with 10--20 B parameters may have $100X$ fewer parameters than a trillion-parameter frontier model. Techniques such as distillation, retrieval, quantization, sparsity, conditional computation, and specialized accelerators provide an alternate path toward  $1000X$ potential reductions in energy per useful task. The rest of the paper presents algorithmic and hardware approaches to achieve this goal.

Achieving brain-like intelligence will require integrated cross-layer co-design of next-generation cognitive systems encompassing novel brain computational models, learning representations and algorithms, and underlying energy-efficient hardware \cite{yirancasm2025}. It took about a decade to evolve from AlexNet with 60 million parameters to GPT-3 with 175 billion parameters, using brute-force training with large datasets. However, truly cognitive systems are in their infancy.
Rapid breakthroughs are needed to achieve true brain-like intelligence 
relative to the state of the art, in ways similar to today's ChatGPT or GPT-4 
compared to the AlexNet of 2012. Between 2013 to 2019, the compute complexity of training AI systems from AlexNet to AlphaGo Zero increased by 300,000$\times$ \cite{synched18}, and is now doubling every two months. The energy consumption of frontier models continues to grow exponentially as well. For example, GPT-4 requires 1.76T model parameters, has been trained using 13T tokens, and consumes 50-60 GWh of energy (about \$63M) for training \cite{thedecoder_gpt4_leak}.  By 2028, the total U.S. data center electricity consumption is projected to roughly double, reaching about 325–580 TWh, or approximately 6.7\%–12\% of U.S. electricity demand \cite{doe2024_datacenters}. This is unsustainable. To reduce the model complexity and improve energy efficiency, collaborative domain-specific cognitive systems need to be developed that can sense, think, react, and collaborate with other cognitive agents specialized in their own domains, to achieve higher-level goals in the real world. This is quite similar to the way the brain fuses information from multiple sensory organs and, based on prior knowledge, experience, and cause-effect reasoning, produces actionable decisions for the future.

Broadly, intelligent systems are defined as ``systems that demonstrate broad capabilities of intelligence, including reasoning, planning, and the ability to learn from experience, and with these capabilities at or above human level" \cite{bubeck2023sparks}. In this context, the field of artificial intelligence has been dominated recently by LLMs that exhibit sparks of intelligence. The evolution of LLM has been described using a seven-layer model \cite{BorsungLiang}. However, despite the initial excitement around LLMs and large vision models (LVMs), progress has been unsatisfactory because current LLMs/LVMs often hallucinate \cite{waldo2024gpts}. One of the reasons LLMs/LVMs hallucinate is that they are trained on data of uneven quality scraped from the web. Efforts are afoot to reduce hallucinations through techniques like retrieval augmented generation (RAG) \cite{lewis2020retrieval}, chain-of-thought prompting \cite{wei2022chain}, and tree-of-thought prompting \cite{yao2024tree}. However, such methods are not able to eliminate hallucinations, preventing them from being deployed in critical application domains such as healthcare, surveillance, and robotics, where accuracy is important, and agent hallucinations may place humans in jeopardy. 

\begin{figure}
    \centering
    \includegraphics[width=\linewidth]{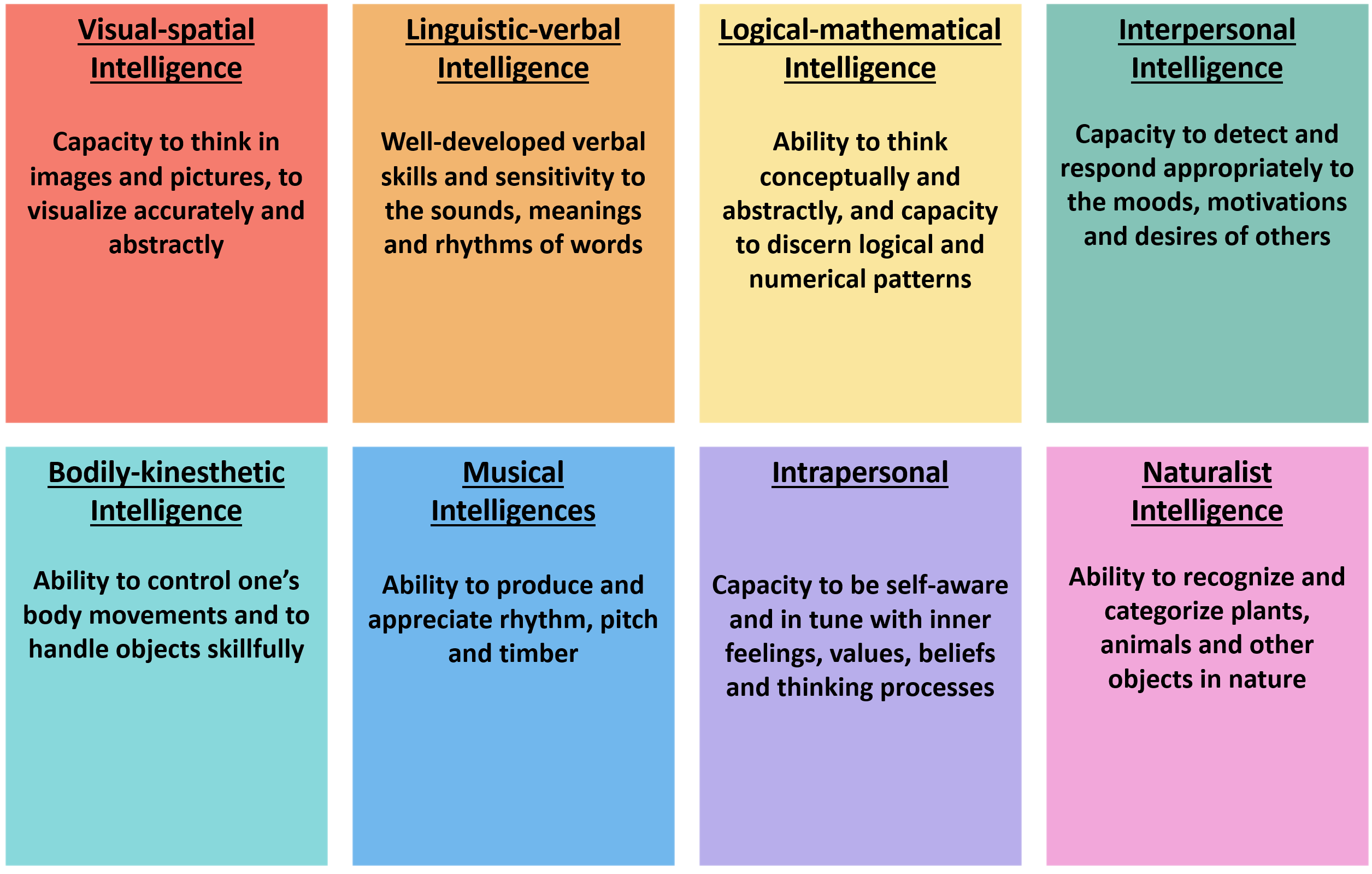}
    \caption{Eight forms of human intelligence proposed by Howard Gardner~\cite{gardner2011frames}. It illustrates complementary cognitive capabilities that can inspire the design of future AI systems and agents.}
    \label{fig:Howard}
\end{figure}

A path forward from this ``hallucination dilemma" is to design AI agents with broader forms of human-like ``general" intelligence.  In this context, Howard Gardner \cite{gardner2011frames} proposed eight different types of intelligences as shown in Fig.~\ref{fig:Howard}: (i) visual-spatial, (ii) linguistic-verbal, (iii) logical-mathematical, (iv) body-kinesthetic, (v) musical, (vi) interpersonal, (vii) intrapersonal, and (viii) naturalistic. Today’s LLMs and Visual Transformers (ViTs) capture aspects of linguistic-verbal and visual-spatial intelligence, including primitive reasoning capabilities
\cite{zhou2023comprehensive,kocon2023chatgpt,touvron2023llama,guo2025deepseek}.  For example, humans or agents with visual-spatial intelligence excel at understanding maps, charts, videos, and pictures; those with linguistic-verbal intelligence excel at summarization, memorization, and storytelling; those with logical-mathematical intelligence excel at reasoning, recognizing patterns, relationships, and logical analysis; and those with interpersonal intelligence excel at understanding the emotions, motivations, desires, and intentions of those around them. Further, these intelligences need to be collaborative and real-time in nature, and feasible on hardware with low energy consumption.

The main contribution of the paper is a unique vision toward domain-specific energy-efficient intelligent systems. The topics of the paper span from reasoning ability requirements to hardware architectures for energy-efficient, domain-specific AI systems. Thus, the paper provides a unified perspective on how domain-specific knowledge representation, efficient computation, and emerging AI architectures jointly contribute to the development of energy-efficient domain-specific AI systems.

This paper is organized as follows. Section~\ref{sec: Capability} identifies the core learning and reasoning capabilities needed for domain-specific models on specialized applications. Section~\ref{sec: Novel-Com} explores novel computational approaches that enable these capabilities with lower computational and energy costs. Section~\ref{sec: Architecture} discusses the emerging AI architectures that integrate reasoning, memory, multimodal perception, and agentic behavior into deployable systems. 

\section{Learning and Reasoning Capabilities for Domain-Specific Systems}~\label{sec: Capability}

This section identifies learning and reasoning mechanisms that enhance the generalization of AI systems from limited data, adapt under distribution shift, and decide when additional computation is worth its energy cost.
We define general intelligence modeled on human brain-like cognition as the ability to handle tasks that differ from previously encountered tasks \cite{chollet2019measure} at two levels of generalization: (a) \textit{local generalization} refers to the ability to generalize across \textit{unseen data from a known distribution} (known-unknown) and (b) \textit{broad generalization} refers to the ability to generalize across \textit{unseen data from an unknown distribution} (unknown-unknown). A key research frontier is that of \textit{developing foundations for broad generalization \cite{chollet2019measure,venkatasubramanian2021towards,hernandez2021general,a100} using minimum amounts of additional data}.
\begin{figure}
	\centering
	\includegraphics[width=\linewidth]{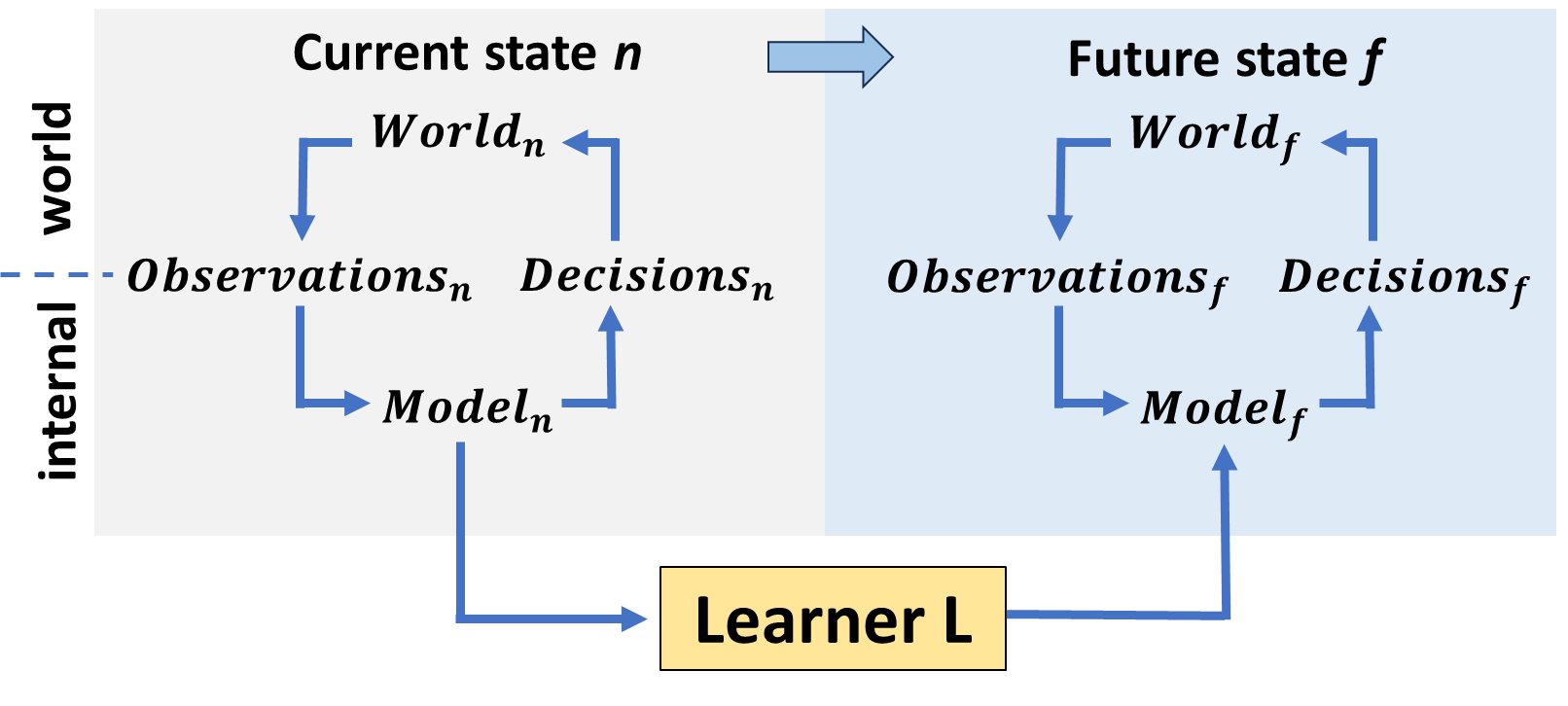}
	\caption{General intelligence, adapted from \cite{vogelstein2022prospective}.}
	\label{world_mod}
\end{figure}

This is illustrated in Fig.~\ref{world_mod} and adapted from \cite{vogelstein2022prospective}. Here, ${World}_n$ and ${World}_f$ are the current and future worlds, ${Model}_n$ and ${Model}_f$ are current and future internal models of the world (dynamically evolving), ${Observations}_{n,f}$ are observations of the world in the current $n$ and future $f$ states of the system, and similarly for decisions ${Decisions}_{n,f}$ as a consequence of those observations.
Learner $L$ learns continuously and updates ${Model}_n$ to reflect the most accurate internal model of the future (unknown) world in ${Model}_f$. The term \textit{future} also implies novel tasks, domain shifts, and unforeseen operating environments based on the application to which generalization is applied.

\begin{figure}
	\centering
    \vspace*{-6mm}
	\includegraphics[width=\linewidth]{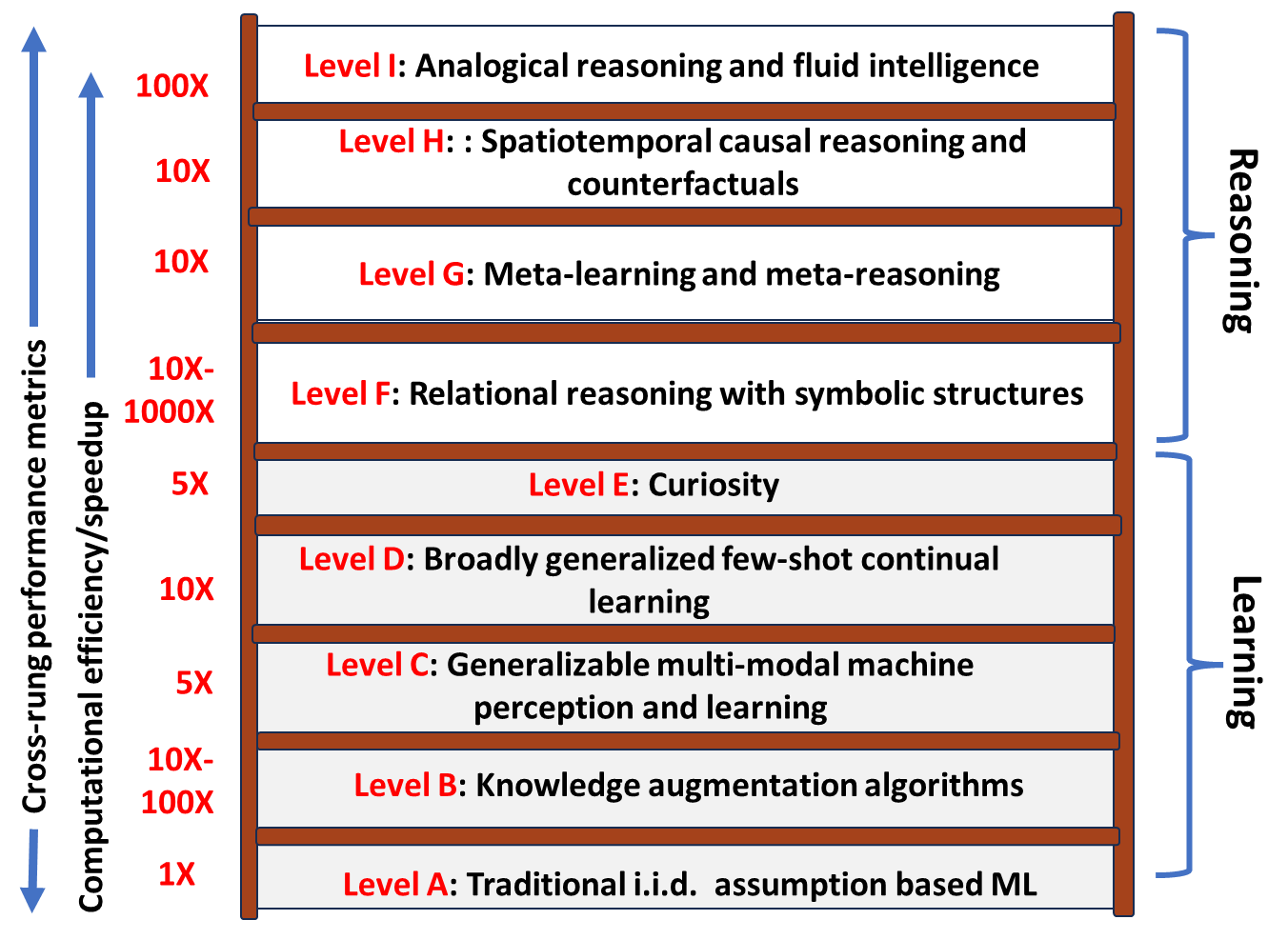}
        \vspace{-0.25in}
	\caption{Ladder of learning and reasoning. Each rung represents an increase in learning and reasoning capability. The red numbers indicate the target system-level reduction in energy per task related to the corresponding level.}
	\label{ladder}
     \vspace{-0.16in}
\end{figure}

Inspired by the work of \cite{pearl2010scm}, this paper proposes a \textit{ladder of learning and reasoning} (Fig.~\ref{ladder}) as the organizing framework.  Levels A-E of the ladder relate to learning paradigms in artificial intelligence, while levels F-I of the ladder relate to reasoning paradigms.
Its lowest rung  
consists of traditional correlation-based machine learning (A).
The rungs above yield collective intelligence from
knowledge augmentation algorithms 
(B), e.g., for foundation models that facilitate continual learning without catastrophic forgetting; generalizable multimodal machine perception and learning (C); broadly generalized few-shot continual learning (D); models for curiosity (E); cognitive inductive biases, constraints, and relational reasoning with symbolic structures (F); meta-learning and meta-reasoning  
(G); spatiotemporal causal reasoning and counterfactuals (H), and analogical reasoning and fluid intelligence (I) placed at the highest rung of the ladder. 
We broadly conjecture that, as we move from the lower to higher rungs of this ladder,
higher levels of generalization of the intelligence of the Learner \textit{\textbf{L}} of Fig.~\ref{world_mod} to future worlds ${World}_f$ are possible. The \textit{reasoning capabilities are leveraged for planning and decision-making} in unforeseen worlds \textit{without access to vast amounts of data, with minimal additional computations}, thereby \textit{increasing computational efficiency}. Fig.~\ref{ladder} also shows expected energy benefits as illustrative target ranges for energy reduction from techniques at each level of the ladder related to the state of the art, running on GPUs. These values are interpreted as system-level designed targets for every workload.
The combined use of these techniques can, ideally, provide overall multiplicative energy benefits, but only when accuracy, latency, and domain-coverage constraints are met. Metrics used to assess intelligence at different levels of the ladder of Fig. \ref{ladder} are predicated 
on existing AI metrics, including Precision, Recall, F-measure, Area-under-curve (AUC), and Accuracy; for natural language processing, metrics such as BLEU score, ROUGE metric, and Word Error Rate; for computer vision, metrics such as Frechet Inception Distance and Structural Similarity Index \cite{blagec2020critical}.  At higher rungs of the ladder,  
statistical tests established in diverse fields of science, finance, and economics can be used.

\begin{figure}
	\centering
    \includegraphics[width=0.8\linewidth]{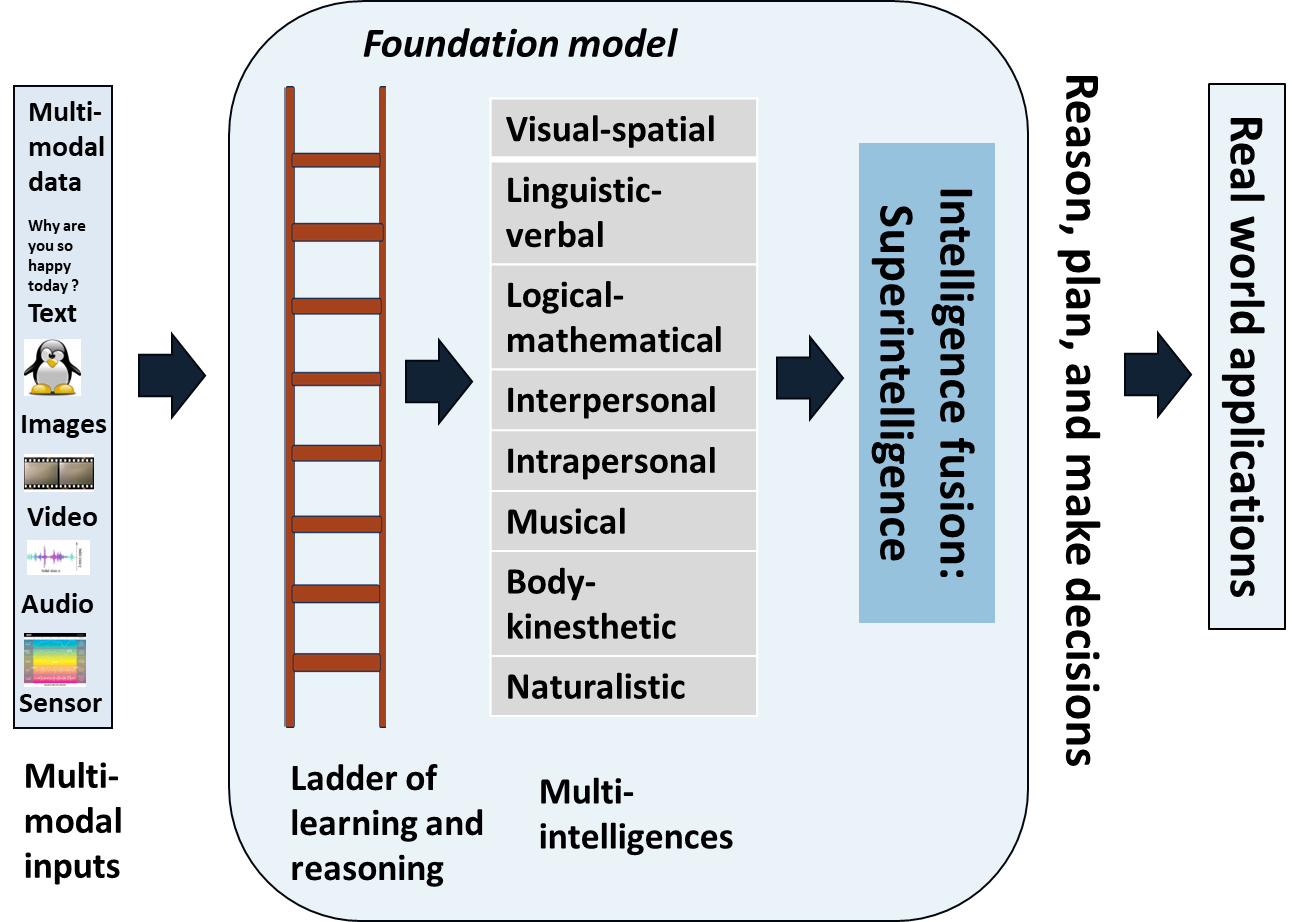}
	\caption{Foundation model for general intelligence.}
	\label{fmgi}
    \vspace{-5mm}
\end{figure}

Fig. \ref{fmgi} shows a foundational model for general intelligence encompassing the ladder of learning and reasoning of Fig. \ref{ladder} and all eight intelligences developed in  \cite{gardner2011frames}. In this roadmap, the figure is treated as a map of capabilities that domain-specific agents may need to combine, in contrast to agents that attempt to achieve general-intelligence. The ladder of intelligence can be applied to each of the eight intelligences of Fig. \ref{fmgi}. The intelligences are then fused together using higher levels of superintelligence to reason, plan, and make decisions with multimodal inputs targeting real-world applications in robotics, transportation, medicine and health, and finance, among others.

Despite the tremendous strides made in AI,
there is still no system that matches the ``sweet spot” of 
flexibility and efficiency exhibited by the human brain
\cite{lake2017building}.
While traditional symbolic processing systems exhibit full
flexibility, they can require laborious programming and cannot match the efficiency of neural networks (NNs) for
many complex tasks, e.g., computer vision or NLP. Conversely, while NN architectures can be trained to perform such tasks efficiently, they require massive amounts of data, and rely on domain-specific function 
approximation that generalizes poorly. 
In contrast, the uniquely human balance of flexibility and efficiency seems 
to reflect the ability to learn abstract, low-dimensional, representations that can be used flexibly for 
inference, reasoning, and generalization across a wide range of domains, while at the same time exhibiting 
the ability to learn task-dedicated, domain-specific forms of processing when performance efficiency is required. 
In the following, we describe the learning and reasoning paradigms at different rungs of the ladder of Fig. \ref{ladder}, starting with analogical reasoning. 

\subsection{Analogical Reasoning for Structural Reuse}

When preschoolers, without any special training, are shown a picture of a tree and asked the counterfactual question, “If a tree had a knee, where would it be?” They typically provide systematic and plausible answers by pointing to the correct area of a picture, even when the task was made difficult by showing the picture upside down or by adding other distractor cues in the picture. Children’s early ability to find correspondences between previously-unrelated concepts demonstrates the creative nature of human intelligence \cite{gentner1977children}. 
This representative example illustrates the synergy of core processes underlying general intelligence. These processes include \textit{relation processing} (e.g., part-whole relations that define body structure), \textit{analogical transfer across domains} (transfer of a known model about body structure to a new object), and \textit{counterfactual reasoning} (imagining a “knee” for a tree). Importantly, children perform this novel task without explicit objectives or any task-specific training. This flexibility contrasts with popular current AI models (e.g., relation network \cite{santoro2017simple}), which depend on direct training on a reasoning task using big data. 

This deep learning approach has been applied with some success to solving visual analogy problems, notably problems consisting of geometric shapes inspired by Raven’s Progressive Matrices (RPM) \cite{raven1998raven}, which are widely used in IQ tests to measure human fluid intelligence (i.e., the ability to reason about novel problems). After extensive training with RPM-like problems, deep neural networks have achieved human-level performance on test problems with a similar basic structure \cite{santoro2017simple}. However, the success of these deep learning models depends on high similarity between training problems and test problems, and on datasets of massive numbers of RPM-like problems (e.g., 1.42 million RPM-like problems); and performance success is defined with a very limited scope (solving RPM-like problems). In contrast, in order for the RPM task to be of interest as a measure of human-like fluid intelligence, extensive pretraining on RPM-like problems must necessarily be avoided. When the RPM task is administered to a person, “training” is limited to general task instructions. 

Despite an upsurge in AI work investigating reasoning and analogy while being grounded in raw inputs, these models are inadequate as algorithms and architectures for general intelligence, as they are unable to achieve human-level generalization in reasoning. The inadequacy is largely due to their dependency on task-specific training, coupled with the assumption that training and testing data consist of independent and identically distributed (i.i.d.) samples from an unknown probability distribution.
What is required in order to capture general intelligence? Psychologists have long referred to two different types of intelligence, each supported by different computational architectures and algorithms \cite{cattell1971abilities}. Crystallized intelligence refers to the accumulation of knowledge, facts, and skills that are acquired throughout life, whereas fluid intelligence refers to the ability to reason and think flexibly. The two types of intelligence have distinctively different characteristics. Crystallized intelligence is rooted in learning experience that yields facts and general knowledge, typically in a verbal format. This type of intelligence becomes stronger as we age, as increasing learning experience allows us to encode new knowledge in long-term memory. In contrast, fluid intelligence is based on the ability to manipulate novel information in working memory in order to solve problems and reason about novel situations. Fluid intelligence tends to increase over cognitive development and then decline during late adulthood.

From a computational perspective, crystallized intelligence could be realized by offline learning with big data across lifelong experience, based on the classical von Neumann processing approach of separating memory and processing. In contrast, fluid intelligence requires more involvement of process-in-memory, flexible information representation and processing, and online computation using a small amount of data. It is well-known that humans have limited capacity in working memory. These limits may contribute to energy efficiency in the human brain, as thinking processes are bounded with limited computational resources at any given time. Because fluid intelligence heavily depends on working memory, the limited capacity of human hardware (i.e., working memory) entails use of novel representations based on selecting, combining, and coordinating input information in a flexible way for reasoning and thinking.

Analogical reasoning allows compact models to reuse relational structure instead of learning each task independently. At a lower level, contrast operations can transform entity representations into relation representations, such as category-membership, part-whole, and cause-effect relations. At a higher level, analogy enables these relations to transfer across domains, which supports zero-shot or one-shot generalization and reduces the amount of training data required for new tasks.
As exemplars of this work, BART (Bayesian Analogy with Relational Transformations \cite{lu2019emergence}) is a learning model for semantic relations. BART takes word embeddings for a small number of data as inputs (e.g., ~20 pairs of words) to learn semantic relations represented with disentangled dimensions in a transformed space. The model builds on contrast operations to implement a module for feature augmentation and feature selection, and estimates associated weights for a relation instantiated by training word pairs. After learning a set of specific relations, BART can compute a high-dimensional vector of relation for any pair of words by calculating the posterior probability that the pair instantiates each of the learned relations. By comparing the similarity between relation vectors, semantic relation representations derived by BART have been used to solve verbal analogies in the A:B :: C:D format \cite{lu2019emergence}, to predict human judgments of relation typicality and similarity \cite{ichien2021predicting}, and to predict patterns of similarity in neural responses to relations \cite{chiang2021distributed}. Simulation results confirm that the contrast-based module for feature augmentation and selection plays an essential role in enabling relation learning with a small number of training data. Recent work enables BART to complete a generative task, such as “Robin is a kind of ?”. Simulation results show that BART with explicit relation representations outperforms transformer models (e.g., BERT) that are directly trained on this type of completion task. This surprising finding points to the importance of explicitly representing relations in reasoning tasks.

A second line of work focuses on developing an inference model, PAM (Probabilistic Analogical Mapping \cite{lu2022probabilistic}), to solve analogies based on more complex knowledge involving several concepts and their interrelationships. The PAM model re-represents stories (e.g., one story is about solar system and the other story is about atom system) as semantic relation networks to compare the two analogs. Each such network is an attributed graph in which nodes and edges are assigned numerical values (attributes) that capture the semantic meanings of individual concepts and their pairwise relations. Using the semantic relation networks created for source and target analogs, PAM performs analogical mapping using a probabilistic approach that jointly maximizes the similarity between the two analogs, with the further constraint of favoring one-to-one mappings between concepts across analogs.  Initial work \cite{lu2022probabilistic} has shown that the PAM model is able to solve complex analogical mappings based on verbal materials. For example, it finds the seven mappings between keywords associated with the Rutherford analogy between the solar system and atom system from raw text inputs. Hence, by building a reasoning model on top of learning mechanisms grounded in distributional semantics, the PAM model has drawn closer to the goal of automating analogical reasoning for natural-language inputs.

\subsection{Prospective Learning, Spatiotemporal Causal and Counterfactual Reasoning for Domain Shift}
Machine learning models today rely on the assumption that the data they have to make predictions on are drawn from the same probability distribution as the data they were trained on. Vogelstein et al. ~call this 
retrospective learning \cite{vogelstein2022prospective}.  However, natural intelligence (that of humans and many
animals) has the ability to learn for a future full of uncertainty. This is called prospective learning
\cite{vogelstein2022prospective}. This would be key to paving the way for artificial general intelligence. Prospective learning has four factors: continual learning, prospective constraints, curiosity, and causal estimation. Continual learning enables the model to retain the most important characteristics of the past that would prove to be fruitful for learning in the future. Constraints, through biases and priors, enable generalization. Curiosity
enables the model to acquire relevant information that may be useful in the future, not necessarily for immediate reward. Causal estimation enables the model to find relationships that persist as opposed to correlations that are context-sensitive \cite{vogelstein2022prospective}. A key problem is that of formulating how knowledge representations stemming from all aspects of prospective learning are stored for efficient re-use. Inter-relationships between knowledge representations at different levels of knowledge granularity need to be stored and indexed to facilitate fast knowledge access using indexing keys containing context and knowledge semantics. Such representations will be hierarchical in nature in terms
of short-term vs.~long-term  (or higher-level) knowledge needs. Prospective learning represents a significant departure from algorithms and architectures that support traditional machine learning, i.e., retrospective learning. Although prior work has targeted the four factors individually: continual training (through incremental training), constraints (to enable regularization and edge implementations), curiosity (in a limited way through exploration), and causality (through interventions and counterfactual reasoning), prospective learning not only needs to simultaneously solve the four sub-problems, it also needs to solve them in a new way. For example, incremental training of a model still assumes that the future looks like the past, i.e., the future and past data are drawn from the same distribution. However, in
prospective learning, we have to discard this assumption, making the problem much harder. Exploitation vs.~exploration techniques are widely used in reinforcement learning. However, while exploitation is greedy, even exploration-based actions are geared towards a near-term reward. This is different from curiosity that is geared towards a reward that may be received in the distant future.

\noindent\textit{Continual learning:} 
A continual learning approach for the 
retrospective learning scenario where a machine learning model is continually updated, but assumes the newly arrived data are drawn from the same distribution, is described in\cite{dai2020incremental}.  It mimics the way neural connections in the human brain 
grow from the baby years to the toddler years and then get pruned in the adult years.  For prospective learning, we discard the assumption that the newly arrived data are drawn from the same distribution. However, we retain the synthesis approach. Neural networks are known to encapsulate new concepts they encounter in polytopes. This may be 
one reason why dynamic inference techniques \cite{xia2021fully}, which allow a convolutional neural network (CNN) to dynamically skip many of its layers at inference time based on the particular received image, have found success.  To accommodate novel data with diverging statistics, the  fixed architecture, substantial training cost, and significant model redundancy of current deep neural network (DNN) architectures pose significant challenges. To solve these problems,
an incremental learning framework based on a grow-and-prune neural network synthesis
paradigm is developed in \cite{dai2019grow,dai2019nest,dai2020incremental,hassantabar2019steerage}. When new data arrive, the neural network first grows new connections based on the gradients to
increase the network capacity to accommodate new data. Then, the framework iteratively prunes away connections
based on the magnitude of weights to enhance network compactness, and hence recover efficiency.
Finally, the model rests at a lightweight DNN that is both ready for inference and suitable for future grow-and-
prune updates. This prior framework has been reported to improve accuracy, shrink network size, and significantly
reduce the additional training cost for incoming data compared to conventional approaches, such as training
from scratch and network fine-tuning.

\begin{figure*}[!htp]\centering
\vspace* {-5mm}
\includegraphics[width = 0.9\textwidth, height = 6cm]{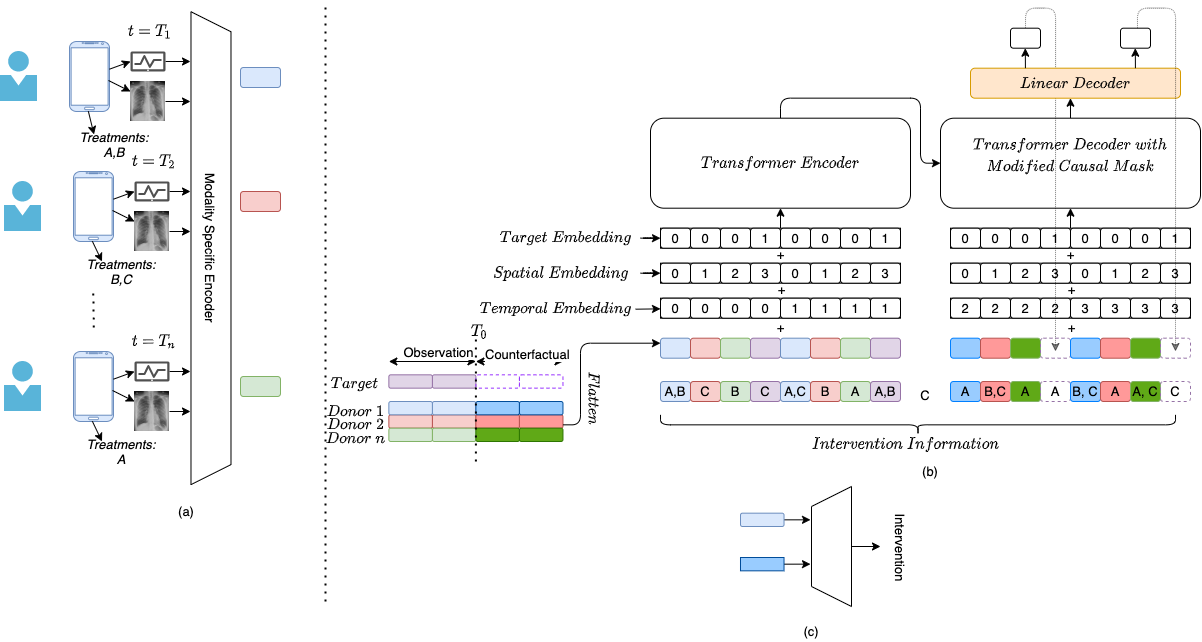}
\vspace* {-2mm}
\caption {Decision-making transformer~\cite{dedhia2023scout}: (a) Modality-specific encoders take input from pocket devices and distill it into a representation, (b) a world model forms the core of the system that learns to predict counterfactuals under various interventions (to do so, it encodes information from donor trajectories and produces outcomes of the target unit under different interventions), (c) an inverse model learns to predict the intervention from representation obtained prior to and after intervention onset (this helps the encoder capture representations that are maximally informative about the counterfactual).}
\label{JhaTransformers}
\vspace* {-4mm}
\end{figure*}

\noindent\textit{Constraints:} 
Machine learning models are often subjected to statistical constraints (to provide an inductive bias or for regularization) and computational constraints (e.g., memory footprint, energy, latency in the case of edge analytics). Both types of constraints are necessary for prospective learning.  
However, the constraints are deployed in a very different way.  In human cognition, constraints imposed on learning are often the result of causal priors.  We are beginning to see the use of causal priors in conjunction with deep learning.  For example, in physics-informed neural networks, one imposes priors from laws of physics to train the neural network
\cite{lavin2021simulation}. This enables the learned weights to be aligned with those laws and, hence, generalize better. However, this approach need not be limited to physics priors. A library of cognitive causal priors can be used to constrain models for deep learning, reinforcement learning, long 
short-term memories, transformers, etc., targeted at various cognitive applications.

\noindent\textit{Curiosity and Curiosity-based Reinforcement Learning:}  In prospective learning, curiosity is directed at maximizing relevant information whereas exploration in reinforcement learning is directed at maximizing current rewards.  In natural intelligence, curiosity is aided by compositional representations, causal relations, and other data/time-invariant relationships. In active learning, curiosity is known to provide an exponential speedup in sample size convergence guarantees. Collecting relevant information entails training the model with spatial configuration, hierarchical relationships, and contingencies \cite{vogelstein2022prospective}.  Note that the traditional motto of transformers is \textit{attention is all you need} \cite{vaswani2017attention}. Prospective learning indicates that \textit{attention is not all you need.}  Curiosity-driven transformers will need to be trained in a very different way, using causal priors as mentioned above, rather than just with a large corpus of text or images. 

\begin{figure}[t]
    \centering
    \includegraphics[width=\linewidth]{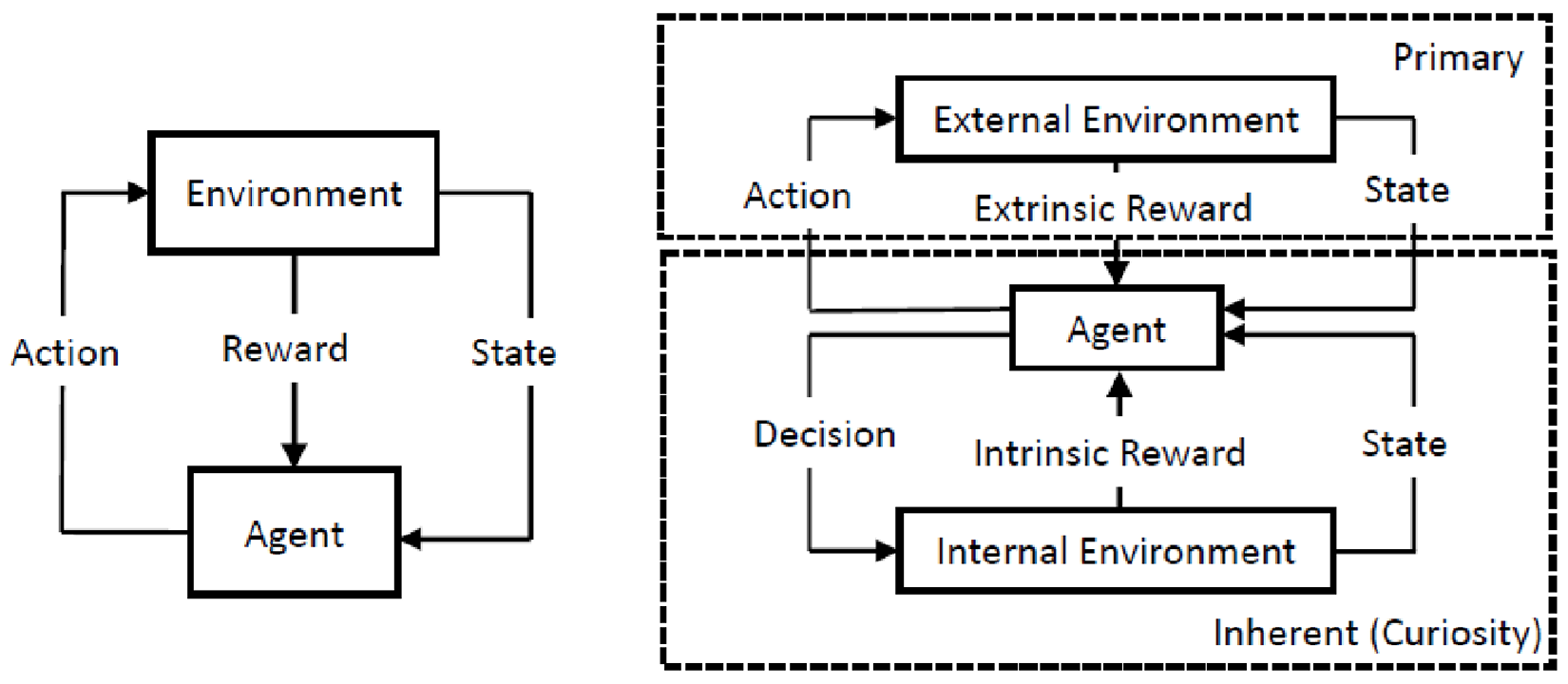}
    \caption{Curiosity-based reinforcement learning~\cite{sun2022psychological}. The agent interacts with the external environment and receives extrinsic rewards. An internal environment or world model predicts outcomes of candidate actions. Prediction errors generate intrinsic rewards that drive exploratory behavior and the acquisition of informative experience.}
    \label{fig:curiosity}
\end{figure}

Curiosity can be operationalized within reinforcement learning by augmenting the standard reward signal with intrinsic rewards that reflect novelty, surprise, or prediction error~\cite{sun2022psychological}. As shown in Fig.~\ref{fig:curiosity}, the agent interacts with the external environment and receives extrinsic rewards. In parallel, an internal environment predicts the consequences of candidate actions. When observations differ from predictions, the resulting curiosity (intrinsic reward) encourages the agent to explore uncertain regions of the state space, which leads to more informative experience and improved future generalization.

\noindent\textit{Causality:} 
Machine learning is built on identifying correlations, which are context-dependent. However, causal relations are data-independent and universally applicable. Despite valiant attempts, teasing causality from observational data has not been shown to be scalable. The gold standard for establishing causality is randomized controlled trials (RCTs). These trials involve an intervention (placement in a control or treatment group) and ascertaining the treatment effect. One of the 2021 Economics Nobel Prize winners was David Card, who investigated natural RCTs (such as a minimum wage increase in New Jersey, but not in neighboring Pennsylvania, to tease out its effect on employment).  Abadie \cite{abadie2021using} pioneered the synthetic control method as a class of powerful
data-driven techniques to estimate the counterfactual reality of a treated unit from several unexposed donor units. At its core, the technique involves a linear model fitted on the pre-intervention period that combines donor covariates to yield the control. However, combining spatial information at each temporal instance using time-agnostic weights fails to capture important inter-unit and intra-unit temporal contexts. A transformer model that leverages particular positional embeddings and a modified decoder attention mask for performing spatiotemporal sequence-to-sequence modeling is more effective for this purpose \cite{dedhia2023scout} (see Fig. \ref{JhaTransformers}). 
The inverse of synthetic control is synthetic intervention \cite{agarwal2020synthetic}.  In this case, we have several treated units and an unexposed donor unit.  The aim is to estimate the counterfactual reality of the donor unit if treated with the same intervention.  In \cite{agarwal2020synthetic}, the synthetic control method of Abadie is generalized to synthetic intervention by again using a time-agnostic linear model.  Thus, it inherits the demerits of the Abadie method in ignoring the temporal context. This can also be rectified using a transformer model.

\subsection{Metareasoning for Adaptive Computation}
Humans have limited cognitive resources and yet are capable of performing an extremely wide range of tasks.  In order to do so, they have to use their limited resources efficiently. This requires identifying effective heuristic strategies, decomposing complex problems into solvable parts, and recognizing when partial solutions can be reused.  By understanding how humans do this, we can develop computer systems that are more efficient, and exploit modularity in task structure, and can solve a wider and more heterogeneous set of problems. Recent work in cognitive science has made significant progress in understanding the efficiency of human cognition by drawing on the formal idea of \textit{rational metareasoning} \cite{lieder2014algorithm,Milli2017,callaway2018learning,callaway2022rational,lieder2017automatic,Lieder2012}. This idea, originally developed in the AI literature, provides a way of formalizing the problem of efficiently allocating computational resources \cite{RussellWefald1991,russell1991principles,russell1997rationality,Horvitz2001Computational}. Imagine an agent faced with a choice between two possible actions.  To evaluate these actions, the agent can execute some computations. For example, the agent could simulate the consequences of selecting one of the options and thus gather information about the reward associated with selecting that action.  However, executing the computations comes with a cost; not just the energy cost of the computation itself, but the time required. Every moment of deliberation is associated with an opportunity cost reflecting the other beneficial actions the agent might have taken in that time.

Formally, we can characterize the {\em value of computation (VOC)} associated with each computation we perform \cite{RussellWefald1991,russell1994provably,Hay2012}. If we are just going to execute a single computation $c$, the VOC of that computation is the expected increase in reward from taking an action $a$ based on beliefs $b'$ informed by $c$ rather than acting on current beliefs $b$, minus the cost involved:
\begin{gather}
   \nonumber\text{VOC}(c,b) = \mathbb{E}_{p(b' \mid b, c)} \left [ 
    \max_{a'} \mathbb{E}\left [ U(a')|b' \right ] - \max_a \mathbb{E} \left [ U(a)|b \right ] 
\right ] \\
- \mbox{cost}(c)
\end{gather}
where $U(a)$ is the (expected) reward associated with action $a$ and $\mbox{cost}(c)$ is the cost function.  Choosing which computation to execute is reasoning about reasoning, metareasoning, and this decision-theoretic formulation of the problem makes it possible to define a rational solution (i.e., execute the action with the highest VOC, unless that VOC is below $0$ in which case act based on current beliefs).

However, it is rare that we are able to solve a problem by executing a single computation. In the general case, when we execute a sequence of computations, calculating the VOC is much more challenging because the outcome of a computation can influence what computation we might perform next.  In this case, we can benefit from formulating the task of selecting what computations to perform as a sequential decision problem: selecting the optimal sequence of computations to perform. This way of expressing the problem has the virtue of making a direct connection to the well-understood framework of Markov decision processes (MDPs). Formally, we can define a meta-level MDP as an MDP where the state space corresponds to the beliefs of the agent, actions correspond to computations that update those beliefs (with a special action that says to stop computing and commit to an action in the external world) \cite{Hay2012}. Rewards reflect both the benefits of the external actions selected by this process and the costs of the computations executed in order to do so.

Meta-level MDPs have been used successfully to model human allocation of cognitive resources in a variety of tasks, including decision-making \cite{lieder2017automatic} and planning \cite{callaway2022rational}. By developing better algorithms and hardware for solving meta-level MDPs, we can transfer the insights from these models to applications where computational efficiency is critical. This formalism can also be used to better design systems for adaptively using computation in deep learning and to develop systems that are able to efficiently reuse past computations  in a modular fashion. What is needed is to 
create systems that approach and exceed the human ability to efficiently use computation leveraging,via (a) designing better solvers for meta-level MDPs, (b) applying this approach to deep learning, and (c) developing systems that can adaptively reuse computations.

\subsection{Relational Symbolic Structures for Compact Abstraction}
Longstanding work on DNNs has established their ability to learn efficient solutions to complex problems, in ways that have often been found to correspond closely to the forms of computation and representation observed in modality-specific areas of neocortex (e.g., occipital \cite{yamins2016using}, parietal \cite{zipser1988back}, medial temporal \cite{banino2018vector}), and that parallel human semantic knowledge \cite{mcclelland2003parallel}. Building on such work, considerable progress has been made in implementing mechanisms in neural architectures that correspond to the attentional, working memory, and cognitive control functions of the human brain. These include the use of recurrent neural networks (RNNs) and long short-term memory (LSTMs) to implement mechanisms for working memory (RNNs \cite{zipser1993spiking}, LSTMs \cite{hochreiter1997long}), together with attentional mechanisms \cite{cohen1990control,vaswani2017attention} that support context sensitivity and sequential processing. These have been inspired by, and used to model the functions of prefrontal cortex and basal ganglia \cite{miller2001integrative,frank2001interactions}. Furthermore, work on complementary forms of learning (rapid episodic; slower semantic; \cite{mcclelland1995there}; \cite{kumaran2016learning}), implemented in neural architectures by augmenting standard DNNs with external memory, have been used to reproduce symbol-like, rule-based forms of computation \cite{graves2014neural,pritzel2017neural}. State-of-the-art modeling efforts implement these components in various combinations, and have broken new ground in approximating human-like performance (e.g., GPT-3, Dall$\cdot$E). However, these still fall short of human capabilities in several ways: a) they require massive amounts of training (i.e., that do not exhibit the data-efficiency of humans); b) they have not yet been found to explicitly represent the kinds of lower-dimensional structure that can be used compositionally in ways that humans seem to do. These are widely assumed to be necessary as a basis for ``explainable'' AI, as well as for causal inference, reasoning, and planning.
Mounting evidence suggests that forms of regularization, coupled with pre-determined architectural factors, may serve as \textit{inductive biases} that constrain general-purpose learning algorithms to develop abstract, low-dimensional representations. Together with domain-specific forms of learning in current applications of DNNs, these may support the unique combination of flexibility and efficiency achieved by the human brain. We identify three factors that may favor the development and processing of abstract, lower-dimensional representations:  \textit{temporal context normalization} \cite{webb2020learning}, \textit{modularization} of processing \cite{dulberg2021modelling,webb2021emergent}, and the regulation of \textit{shared vs. separated} representations.

\begin{figure}
\centering
	\includegraphics[scale=0.4]{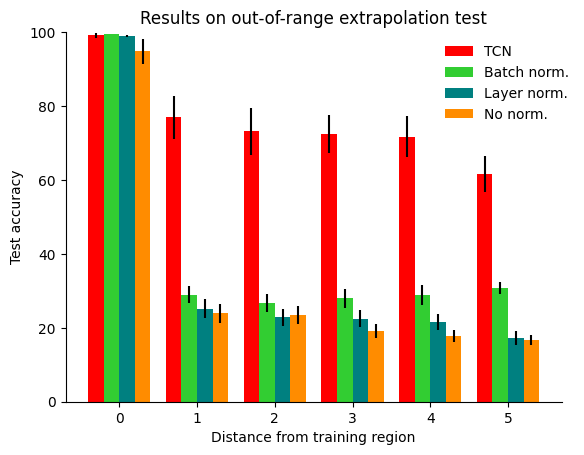}
	\vspace{-1pt}
	\caption{Performance on out-of-range extrapolation (Y axis; chance=14\%), for networks with temporal context normalization (TCN), other common forms of normalization (batch or layer normalization), or no normalization, grouped along X axis for test sets of objects drawn from ranges progressively further away from the training region (0=objects within the training region, i.e., interpolation). Error bars represent the SEM over eight trained networks~\cite{webb2020learning}.}\label{jonathan-cohen-fig:1}
	\vspace{-0.25in}
\end{figure}

\noindent{\textit{Temporal context normalization:}} A key factor in the flexibility of human behavior is the ability to generalize, via interpolation and extrapolation, what has been learned in one domain to others. Interpolation refers to successful processing of items not previously experienced, but within the range of values (``convexity'') of that experience; neural networks are highly effective at this.  Extrapolation is more challenging, involving the ability to generalize \textit{outside the range} of training examples. Humans can often do this, but it has yet to be fully implemented in artificial NNs. For some tasks, traditional computers excel at extrapolation (e.g., the ability to do arithmetic); however, they must be hardwired and/or programmed to do so.  
Several mechanisms have been described for inducing networks to acquire abstract, structured representations that can be used for out-of-range generalization (e.g., \cite{kingma2019introduction,radford2015unsupervised}); however, these tend to require careful parameterization and/or substantial amounts of custom crafted training. We recently identified a simple, more robust mechanism, that uses temporal context normalization, inspired by mechanisms observed widely in the mammalian neocortex \cite{carandini2012normalization}. We demonstrated this in a model trained to perform a novel visual analogy task. The model comprised a standard CNN (for processing images), coupled with an LSTM (for active maintenance, control, and decision-making), and equipped with a mechanism for temporal context normalization of encoded information passed to the control layer. Normalization was applied to the temporal sequence of stimuli (``context frame") that made up each analogy problem (see \cite{webb2020learning} for details). As shown in Fig. \ref{jonathan-cohen-fig:1}, the model was able not only to successfully interpolate within the range of training objects, but also extrapolate to feature values well beyond the range of training, qualitatively outperforming other state-of-the-art models and normalization techniques (e.g., \cite{ba2016layer,ioffe2015batch}). A limitation of this model, however, is that the context frame for temporal normalization was tailored to the task on which it was trained. To apply this mechanism more generally, the system must be able to determine, on its own, the context frame for normalization appropriate for a given task.

\begin{figure}
	\centering
	\includegraphics[scale=0.6]{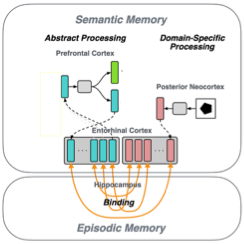}
	\vspace{-1pt}
	\caption{Diagram of Emergent Symbol Binding Network (ESBN) model~\cite{webb2021emergent}, consisting of domain-specific content (pink) and abstract processing (blue) components that interact only via bindings (orange) in episodic memory, allowing the abstract component to learn to represent rules and variables that are present in but abstracted away from the specific content.}\label{jonathan-cohen-fig:2}
	\vspace{-0.25in}
\end{figure}

\noindent{\textit{Modularization and symbol processing:}} While context normalization provides a means of out-of-domain generalization, the full flexibility exhibited by human processing relies not only on the acquisition of abstract representations, but also the ability to flexibly bind them to relevant, task-specific representations. This is often referred to as ``variable binding,'' and is subserved by symbols in traditional architectures. This ability is fundamental to the most abstract human capabilities, such as mathematics, that have eluded artificial NNs \cite{holyoakproper,kriete2013indirection,marcus2001algebraic}.  Some lines of work have proposed explicitly mathematical solutions to this problem (e.g., the use of tensor product representations \cite{plate1995holographic,smolensky1990tensor}), but it is not clear whether or how these might be implemented in the brain, and to what extent they can explain its full range of processing capabilities. Another line of work has explored the integration of neural networks with an episodic memory mechanism for binding \cite{botvinick2019reinforcement}. As discussed below, this has potential neural plausibility.  While previous work using this approach has yet to achieve human levels of data efficiency and flexibility (e.g., \cite{graves2014neural}), recent work has identified a simple augmentation of this approach that promotes the acquisition of abstract, and low-dimensional representations: the separation of processing into distinct streams, one of which makes use of standard mechanisms (e.g., CNNs and LSTMs) to encode and decode domain-specific information, and another that is isolated from the first, but trained using a loss function that is coupled to the first.  The two streams are synchronized when accessing the episodic (``external'') memory mechanism, that serves to bind (rapidly associate) items written to it at the same time (see Fig. \ref{jonathan-cohen-fig:2}), thus linking processing in the two streams. Critically, during learning, this configuration induces the second (abstract) stream to: i) extract abstract relationships (i.e., ``rules'') among items in the first (content) stream that are independent of its domain-specific content; ii) develop generic ``keys'' corresponding to distinct roles in the rules learned by the abstract stream; iii) bind those keys to relevant ``values'' (fillers) in the content stream. Thus, the abstract stream can be thought of as learning a function (set of rules) needed to perform the task, and to bind its keys (symbols corresponding to variables) to the relevant values in the content stream on which the function is being computed. This architecture is inspired by, and consistent with interactions between the prefrontal cortex (abstract stream), the posterior neocortex (content stream), and the hippocampus (episodic memory) in the brain (see Fig. \ref{jonathan-cohen-fig:2}).  Prior studies have shown that a model based on this architecture can perform a suite of abstract rule learning tasks, including a simple version of RPM used widely in intelligence tests. Furthermore, with training on as few as a \textit{single set} of items exemplifying a rule, it can generalize to an arbitrarily large, fully novel set of items, substantially outperforming other state-of-the-art network architectures (including transformers and the Neural Turing Machine). Critically, these results demonstrate not only the learning and generalization capabilities of the network, but that these are achieved by learning a small set of fully abstract representations -- effectively symbols -- that it relies on to perform the task.  The architecture has also been applied to other basic functions, such as counting, that are fundamental to learning more complex capabilities such as mathematical reasoning (e.g., \cite{sarnecka2008counting,schaeffer1974number}), showing not only that the architecture can learn these tasks, but also that it does so with a developmental trajectory that closely follows human acquisition \cite{lee2010model}.

\noindent{\textit{Shared vs. separated representations and cognitive control:}} The flexibility of abstract representations relies on their generality: the ability to share their use across multiple domains. This has been most evident in meta-learning in multi-task training environments, where success relies on the development of representations that can be shared across many task-specific contexts. However, complementary theoretical work has shown that the benefits of such shared-use representations for \textit{flexibility} comes at a cost: constraints on the \textit{efficiency} of processing. Tasks that rely on shared representations must be \textit{serialized} to avoid cross-talk. That is, use of shared representations precludes the efficiency of parallel processing (i.e., multitasking). This can be overcome through the formation of \textit{separated}, task-dedicated representations, that permit parallel execution. However, that comes at the cost of more training and poorer generalization. Prior work has shown that the management of this tradeoff \textemdash{}~between the representational \textit{flexibility} of abstract, shared representations, and the processing \textit{efficiency} of separated representations \textemdash{} provides a normative, formally rigorous, and empirically testable framework for understanding the longstanding distinction between controlled and automatic processing in human performance, and the progression from one to the other with practice \cite{feng2014multitasking,alon2017graph,sagiv2018efficiency,musslick2021rationalizing,petri2021topological}. It also suggests a fundamental way in which the hierarchical organization of NN architectures interacts with representational structure and control: More abstract representations are valuable precisely because they are \textit{shared}; they can be used by a broad range of processes and tasks. However, as a consequence, they come with a requirement for serialization and thus regulation by control. This may explain why, in humans, the most abstract forms of processing (such as mathematical reasoning) also appear to be the most control-dependent and rely so consistently on prefrontal cortex function. This perspective also frames normative questions, such as how the system may seek to maximize future cumulative reward by shaping its representational structure to optimize the tradeoff between the benefits of shared representations (higher immediate but lower asymptotic performance) and those of separated representations (lower immediate but higher asymptotic performance). 

\begin{figure*}
	\centering
	\includegraphics[width=\linewidth]{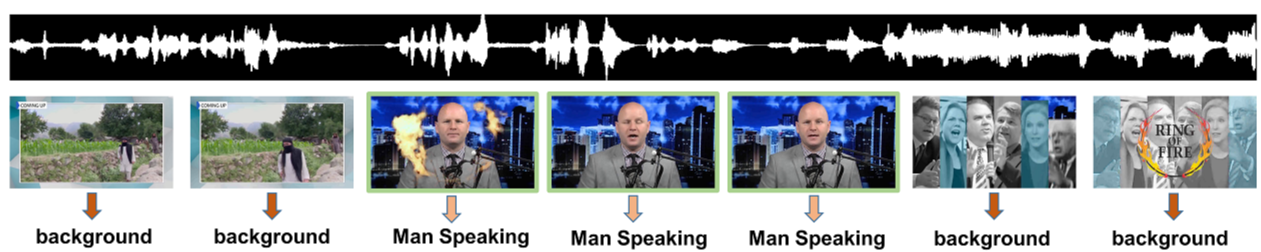}
	\vspace{-1pt}
	\caption{Example of an audio-visual event (AVE) representing the event of an individual speaking. The person’s voice is audible in all of the frames. Only when the person is visible, an AVE is
identified.}\label{aveclip}
	\vspace{-0.25in}
\end{figure*}

Abstract reasoning tasks can be defined as tasks that involve processing sensory information in accordance with abstract rules. These rules are often unknown and must be inferred by AI models. To address this, Ref.~\cite{hersche2023neuro} employs neuro-vector symbolic architectures, which map sensory data into a high-dimensional space. This mapping enables the construction of structures that integrate both sensory and relational information~\cite{hersche2023neuro}, where relational information refers to correlations between individual sensory datapoints. Thanks to the \textit{blessing of compositionality}~\cite{menet2023mimonets}, such structures can be built with minimal interference. Using this approach, Ref.~\cite{hersche2023neuro} successfully solves various RPM tasks.  

However, this method requires prior knowledge and hand-engineering of the atomic features that compose each sensory datapoint, such as the color and shape of objects in RPMs. To overcome this limitation, Ref.~\cite{mejri2024resolve} proposed combining transformers with vector symbolic architectures (VSA). This hybrid approach leverages the power of hyperdimensional structures, which allow for the efficient integration of information with low interference, together with the ability of transformers to automatically extract features from raw input data. Specifically, the sensory features extracted by the transformer are disentangled from the relational information that defines the abstract rule, following the relational bottleneck framework used in \cite{altabaa2023abstractors,webb2021emergent,webb2024relational}. These two structured representations are then recomposed in high-dimensional space. This technique has demonstrated the ability to solve relatively complex sequence-to-sequence abstract reasoning problems \cite{mejri2024resolve}.

While relational and symbolic mechanisms reduce the data needed to learn rules, the useful domain-specific agents are also required to bind those rules to text, images, video, audio, and sensor streams. The next subsection moves from abstract structure to multimodal perception and learning, where the question is how compact agents can perceive domain evidence without paying the full cost of a general-purpose multimodal foundation model.

\subsection{Multimodal Perception and Learning}
Multimodal tasks, especially those relying on generative models (e.g., audio-video event identification, open vocabulary semantic segmentation, etc.), introduce challenges
driven by increased computational costs. CLIP \cite{radford2021learning} learns directly from raw text about images with the goal of predicting an image that is most likely to represent a textual description. This is applied to diverse computer vision datasets encompassing action recognition in videos, geo-localization, and other fine-grained object localization tasks. In addition to the CLIP-image encoder operating on successive video frames, a knowledge distillation-based learning scheme is introduced in \cite{mahmud2023clip4videocap} that aims to exploit the CLIP-text encoder to generate rich textual knowledge from the image features. For improved temporal reasoning over the video, a multi-scale temporal fusion scheme that accumulates temporal features from different temporal windows is developed. In addition, various commonsense aspects in the caption generation are integrated that greatly enhance the caption quality by extracting commonsense features from the video in the intermediate phase. There has also been work on language-vision-audio integration into machine learning frameworks.  For example, an audio-visual event (AVE) is denoted by the correspondence of the visual and auditory signals in a video segment.
Precise localization of the AVEs is very challenging since it demands effective multimodal feature correspondence to ground the short and long-range temporal interactions. Existing
approaches struggle to capture the different scales of multimodal interaction due to ineffective multimodal training strategies. To overcome this limitation, AVE-CLIP \cite{mahmud2023ave}, a novel framework that integrates the AudioCLIP \cite{guzhov2022audioclip} pre-trained on large-scale audio-visual data with a multi-window temporal transformer to effectively operate on different temporal scales of video frames is developed. Fig. \ref{aveclip} shows one instance of the audio-vision event identification problem. Further, a new parameter-efficient
audio-visual transformer employing deep modality alignment for corresponding multimodal semantic features is developed in \cite{mahmud2024ma}. This introduces joint unimodal and multimodal token learning for aligning the two modalities with a frozen modality-shared transformer and allows the model to learn separate representations for each modality, while also attending to the cross-modal relationships between them. In addition, blockwise contrastive learning is used to align coarse-to-fine-grain hierarchical features throughout the encoding phase. Other more complex problems, such as the use of text, visual, and audio inputs for sound source localization in images, are developed in \cite{mahmud2024t}.  In this approach, the textual representation of each sounding source is employed as guidance to disentangle fine-grained audio-visual source correspondence from multi-source mixtures, leveraging the tri-modal Audio-CLIP embedding. This approach enables handling of a flexible number of sources and exhibits promising zero-shot transferability to unseen classes during test time. 

\begin{figure*}
    \centering
    \includegraphics[width=\linewidth]{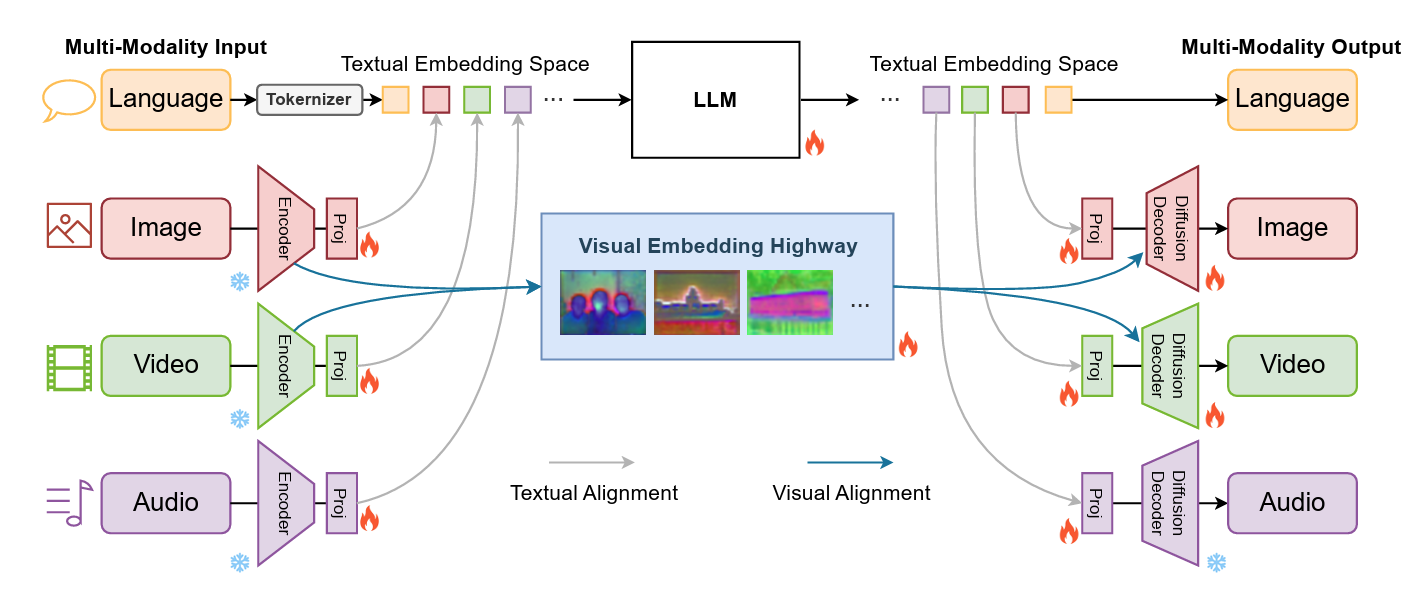}
    \caption{X-VILA schematic diagram~\cite{ye2024xvilacrossmodalityalignmentlarge}. X-VILA augments a pretrained LLM towards new modalities via (i) connecting pretrained encoders to the LLM input textual embedding space and (ii) connecting pretrained diffusion decoders to the LLM output textual embedding space. The system is jointly trained via a new cross-modality alignment procedure.}
    \label{fig:x-vila}
\end{figure*}

An approach to fusion of multimodal information such as text, image, video, and audio,  referred to as X-VILA, has been recently proposed~\cite{ye2024xvilacrossmodalityalignmentlarge}. In this approach, modality-specific encoders are aligned with LLM inputs, and diffusion decoders are aligned with LLM outputs, X-VILA achieves cross-modality understanding, reasoning, and generation. Furthermore, cross-modality alignment is achieved by effectively interleaving any-to-any modality instruction-following dataset. The X-VILA architecture is illustrated in Fig.~\ref{fig:x-vila}.

\subsection{Knowledge Augmentation for External Memory}
Continual learning, learning with constraints, curiosity, and causality generate knowledge that must be organized for reuse rather than stored entirely in model parameters. Such knowledge needs to be well organized and structured 
into hierarchies as per their need for cognition and decision-making: short-term vs.~long term. System 1 knowledge can be stored in NN parametric memory, is gigascale, and consists of instantaneous, intuitive, and habitual knowledge. System 2 knowledge is terascale, stored in an adjacent structured integrated knowledge base and consists of standby information, logic, planning, and reasoning. A third level of knowledge is information from a large repository that is zetascale and contains relevant knowledge from diverse remote information sources \cite{singer22}. There has been significant work on hierarchical structured (e.g., graph-based) knowledge representations \cite{sarrafzadeh2020hierarchical,zhang2020learning,zhang2020relational,webb2021emergent,musslick2019mechanistic,ichien2021predicting}. For domain-specific systems, the relevant future direction is to develop structured knowledge representations that capture the context and semantics of System 1 and System 2 entities and their inter-relationships. This can drive algorithms for knowledge caching using context and semantic information queries. 
This section identifies capabilities that allow compact domain-specific AI systems to rely less on memorized parameters and more on structured reasoning and trusted external knowledge.

\section{Efficient Compute Paradigms for Domain-specific Systems}~\label{sec: Novel-Com}
\begin{figure}
    \centering
    \includegraphics[width=\linewidth]{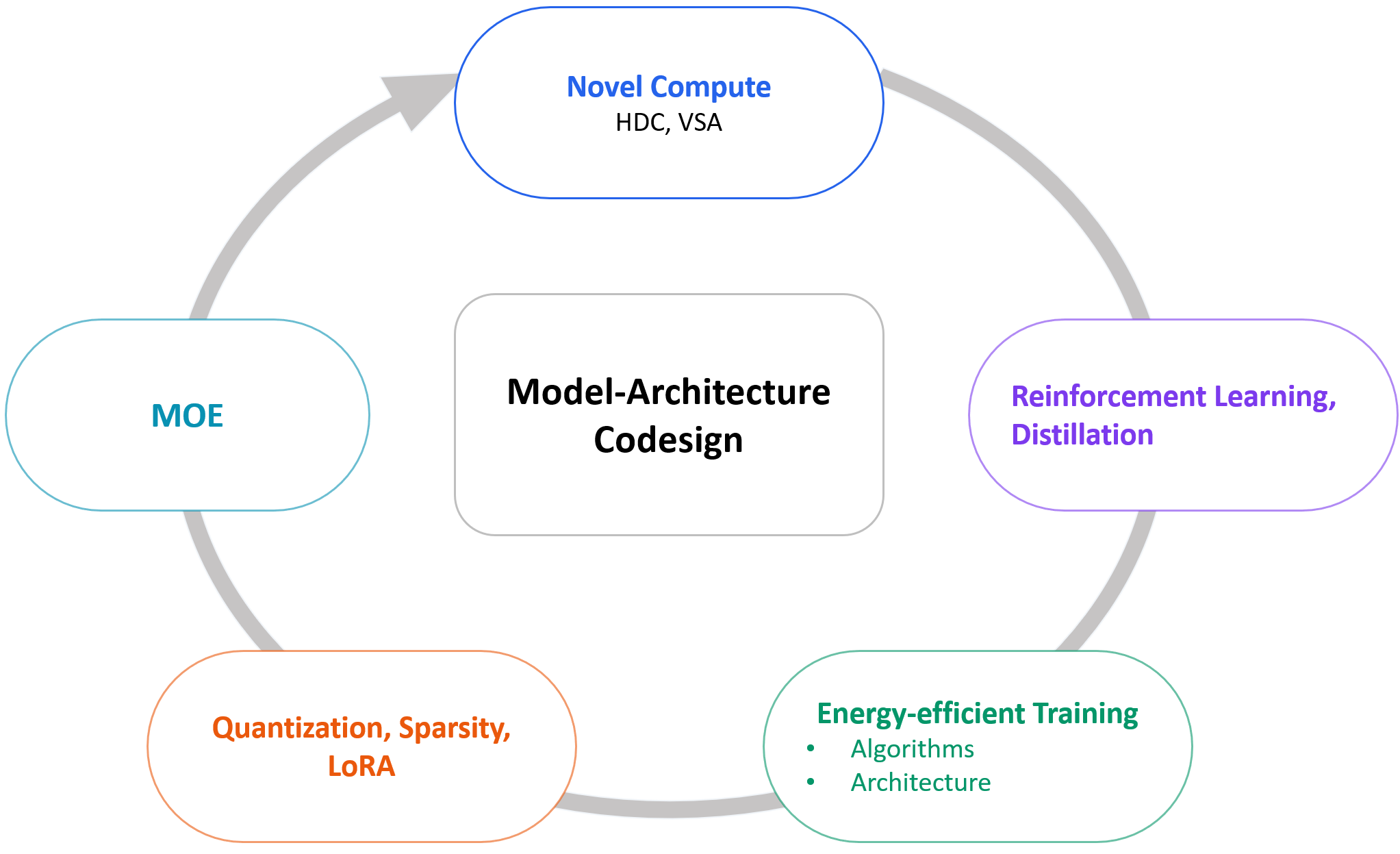}
    \caption{Model--architecture codesign for compute-efficient domain-specific AI systems. Five complementary research directions jointly improve computational efficiency for model-architecture codesign through compact representations, efficient reasoning, optimized training, parameter-efficient adaptation, and conditional computation.}
    \label{fig:codesign}
\end{figure}

This section moves from the learning and reasoning capability that compact domain-specific AI systems are required to have, to the efficient way those capabilities can be executed. The main goal is to reduce the computation and energy cost required for each correct domain-level decision, rather than improving benchmark accuracy. As shown in Fig.~\ref{fig:codesign}, these techniques work jointly to improve efficiency for model--architecture codesign across the entire AI pipeline. Hyperdimensional computing and vector symbolic methods provide compact representations; reinforcement learning and distillation train smaller models to reason effectively; efficient training lowers adaptation cost; quantization, sparsity, and low-rank approximation reduce memory traffic; and mixture-of-experts activates the partial model components needed for a given input.

\subsection{Hyperdimensional Computing for Compact Representation}
Hyperdimensional computing (HDC) is an alternative paradigm of computing based on a different information representation, where all the information is evenly distributed in its unique datatype---\textit{hypervectors}. Such hypervectors are ultra-wide vectors with dimensionality $d$ typically in thousands, e.g., $d=10,000$. In general, HDC allows various types of input data, including letters, signals, and images. Prior work has demonstrated its feasibility for DNA sequencing \cite{imani2018hdna}, speech recognition \cite{imani2017voicehd}, language recognition \cite{joshi2016language}, and biomedical applications using electromyography/electroencephalography~\cite{burrello2020hyperdimensional,rahimi2018efficient}. As a form of brain-inspired computing \cite{parhi2020brain}, HDC continues to draw significant attention due to its comparable accuracy with traditional machine learning models, one-shot learning ability, high energy efficiency, and fast execution time. A comprehensive review of HDC for classification is provided in \cite{ge2020classification}. In the context of seizure detection, three approaches to encode the power spectral density features through HDC are proposed in \cite{ge2021seizure},~\cite{applicability_hdc}. These
approaches are referred to as \textit{single classifier long hypervector}, \textit{multiple classifiers}, and \textit{single classifier short hypervector}. The simulation results show that feature selection plays an important role in seizure detection based on HDC. Selected features using a single classifier short hypervector achieve the optimal detection performance and requires minimum memory resources. 

HDC is an emerging classification approach where data can be classified using one long class hypervector per class. HDC can achieve good accuracy with less training data and is suitable for one-shot or few-shot learning. It can also reduce energy consumption for inference in edge devices. During training, the class hypervector is trained from features extracted from training samples. During testing, the query hypervector is compared with the class hypervector and the label of the class hypervector closest to the query hypervector is selected as the output label. While HDC has been applied successfully in some applications, it fails to demonstrate reasonable performance in other applications and the reason behind HDC's failure in certain tasks is still under study, as well as the relative significance of feature vectors for a specific task. More research work is required to completely understand theoretical foundations of HDC and the goal is to develop a training scheme for online learning with better performance in out-of-distribution data points. 

Future cognitive computing systems require new research in HDC-based learning systems. Future edge computing devices can benefit from low training cost and low energy consumption for inference using a hyperdimensional representation. 
HDC is also interesting from an algorithmic perspective. 
The well-known ``curse'' of dimensionality causes two random vectors in a high-dimensional space to be nearly unrelated (i.e., orthogonal) with high probability. This provides two intriguing properties in cognitive modeling: (1) independent random hypervectors will be unrelated and so can naturally represent objects that are semantically separate, e.g., letters of the alphabet; (2) two hypervectors can be classified as being related (i.e., somehow dependent) with high probability without needing a high inner-product similarity, hence being robust to noise. These properties are similar to what a human brain has; thus HDC is brain-inspired.
HDC is also advantageous from a computer-architectural perspective. In an HDC system, binding, bundling, and permutation are the typical compute primitives. These primary operations are simple arithmetic and have massive parallelism, which project HDC into the scope of energy-efficient and ultra-low-latency computing, especially with the rise of emerging hardware such as processing-in-memory. HDC has recently garnered considerable attention from the point of view of edge applications, e.g., robotics, genomics, health diagnosis, as well as recommendation systems.

Unfortunately, the practical deployment of HDC is undermined by its lower model accuracy compared to alternative methods.
There are two main approaches in the literature to improve HDC performance.
One approach is to increase the hypervector dimension, staying within the classic HDC paradigm and just making the binary vectors longer \cite{neubert2019hdcRobotics,schlegel2021vsa}.
An alternative is to increase the complexity of each element in a hypervector, e.g., to floating-point or complex numbers (unit circle in the complex plane) \cite{plate1995holographic,gallant2013represent,Gayler1998MultiplicativeBR,plate1994complexHDC}:
this moves the system into the more general realm of \emph{vector symbolic architecture} \cite{schlegel2021vsa}, which uses high-dimensional vectors with elements that are not necessarily binary (unlike HDC).

\subsection{Reinforcement Learning Driven Reasoning and Distillation}
Reasoning-oriented reinforcement learning offers a way to train compact systems to selectively spend computation on hard domain questions, rather than increasing parameters indiscriminately. DeepSeek-R1\cite{guo2025deepseek} is the first influential work on introducing reasoning capabilities in LLMs without any supervised finetuning and human feedback. The authors argue that reasoning-oriented group relative policy optimization(GRPO)~\cite{GRPO} can improve reasoning capabilities of LLM models and enable the model to exercise chain-of-thought for solving complex reasoning and mathematical problems. Unlike the prominent proximal policy optimization (PPO)\cite{ppo_paper} scheme, where the critic model is almost comparable in size to the policy model, the DeepSeek-R1 model adopts GRPO\cite{GRPO} to align the language model. Without any value model, the algorithm formulates multiple possible solutions for a given prompt, and each solution is evaluated using a rule-based reward model. The reward of a response consists of two components: the first is an accuracy reward that takes the correctness of the response into account; the second reward component is the format reward, which evaluates whether the model is able to follow a predefined thinking process. For each query $q$, the GRPO\cite{GRPO} generates a bunch of sample responses ${O_1, O_2, O_3,..., O_g}$ from the previous policy model $\pi_{\theta_{old}}$ and the policy function $\pi_{\theta}$ is optimized by maximizing the following objective function:
\begin{equation}
\begin{aligned}
\mathcal{J}_{G R P O}(\theta) =\mathbb{E}\left[q \sim P(Q),\left\{o_{i}\right\}_{i=1}^{G} \sim \pi_{\theta_{\text {old }}}(O \mid q)\right]
\end{aligned}
\end{equation}

\begin{equation}
\begin{aligned}
 \frac{1}{G} \sum_{i=1}^{G}\left(\min \left(\mathcal{R}_{\pi}A_{i}, \operatorname{clip}\left(\mathcal{R}_{\pi}, 1-\varepsilon,1+\varepsilon\right) A_{i}\right)\right)- \beta \mathbb{D}_{K L}
\end{aligned}
\end{equation}
where
\begin{equation}
\mathcal{R}_{\pi} = \frac{\pi_{\theta}\left(o_{i} \mid q\right)}{\pi_{\theta_{\text {old }}}\left(o_{i} \mid q\right)}
\end{equation}
$\epsilon$ and $\beta$ are hyperparameters, and the Kullback-Leibler\cite{KL1,KL2} divergence between the reference policy $\pi_{ref}$ and the new policy $\pi_\theta$ is defined by

\begin{equation}
    \mathbb{D}_{K L}(\pi_\theta||\pi_{ref})=\frac{\pi_{r e f}\left(o_{i} \mid q\right)}{\pi_{\theta}\left(o_{i} \mid q\right)}-\log \frac{\pi_{r e f}\left(o_{i} \mid q\right)}{\pi_{\theta}\left(o_{i} \mid q\right)}-1
\end{equation}

In PPO~\cite{ppo_paper}, the advantage function, $A_i$, is determined from the critic model given the question, but training the critic model is inefficient and complicates the training process. In GRPO~\cite{GRPO}, the relative advantage of individual responses is calculated from the distribution of the rewards within the group of responses the model generated in response to the question. This notion is consistent with the fact that reward models for LLMs are indeed comparing in nature, and advantage factors in this method are defined as: 

\begin{equation}
    A_{i}=\frac{r_{i}-\operatorname{mean}\left(\left\{r_{1}, r_{2}, \cdots, r_{G}\right\}\right)}{\operatorname{std}\left(\left\{r_{1}, r_{2}, \cdots, r_{G}\right\}\right)}
\end{equation}
The DeepSeek-R1-Zero model is observed to demonstrate self-verification, reflection, and the ability to generate a long chain of thought without the need for supervised fine-tuning. The model is able to solve complex reasoning problems by extending the range of chain of thought processes, simultaneously exploring as well as refining the overall thought process. With increased test-time computation, the model is able to achieve complex reasoning abilities with the help of self-evolution.

Furthermore, DeepSeek-R1 demonstrates that the reasoning capabilities of LLMs can be distilled into smaller models, and these dense student models exhibit comparable reasoning performance to large language models that are significantly larger in the number of model parameters. Taking DeepSeek-R1 as the teacher and smaller standard models such as QWEN\cite{qwen1,qwen2} and Llama\cite{touvron2023llama} as student models, supervised fine-tuning is seen to achieve high reasoning performance on small-scale student models. This result supports the main claim that the expensive frontier model can serve as a teacher, while the compact system is a student augmented with retrieval, structured knowledge, and domain-specific verification.

\subsection{Energy-Efficient Training}

DNNs~\cite{parhi2020brain,alexnet,VGG,resnet} account for a significant part of the datacenter workloads and are used for applications such as recommender systems, automated photo recognition, and automatic text generation. Thus, acceleration for on-device inference has been extensively investigated with numerous architectures and variants~\cite{permdnn,eyerissv2_19}. However, as DNN sizes have been continuously increasing, the one-time training cost is no longer insignificant~\cite{Bert,625000CO2}. Furthermore, training with accelerators and distributed heterogeneous training frameworks are still in their infancy. In addition, the larger models can no longer fit within a single processor, necessitating parallelism that requires optimal network partitioning and scheduling. Training optimization must explore a method to maximize the reuse of variables within the system. Existing coarse-grain partition schemes result in an imbalanced distribution of workloads. This is further complicated with heterogeneous systems that lead to low energy efficiency due to processor underutilization. The major bottleneck in terms of energy consumption during training is the backpropagation algorithm. The algorithm computes the gradients of the loss function at each layer with respect to the current layer's weights and the previous layer's activation outputs. A key aspect of optimizing the backpropagation algorithm is understanding the interactions between these operations. 

The fundamental operations in DNN training using backpropagation fall into two categories. The first is common core operations shared between many different network architectures. For example, matrix multiplications, activations, pooling, and softmax operations are common to several DNNs. 
The second is network-specific operations, such as multi-head attention for transformers, aggregation for graphs, and im2col for convolutions.
The backpropagation algorithm consists of three major steps: (i) computing the gradient of the loss function with respect to the output of the activation function ($\boldsymbol{\delta}^{(l-1)}$),
(ii) computing the gradient of the loss function with respect to the weights ($\boldsymbol{W}^{(l)}$),
and (iii) updating the weights.
Traditional approaches treat these computations as independent operations, which is inefficient. 

\begin{figure}
	\centering
	\includegraphics[scale=0.3]{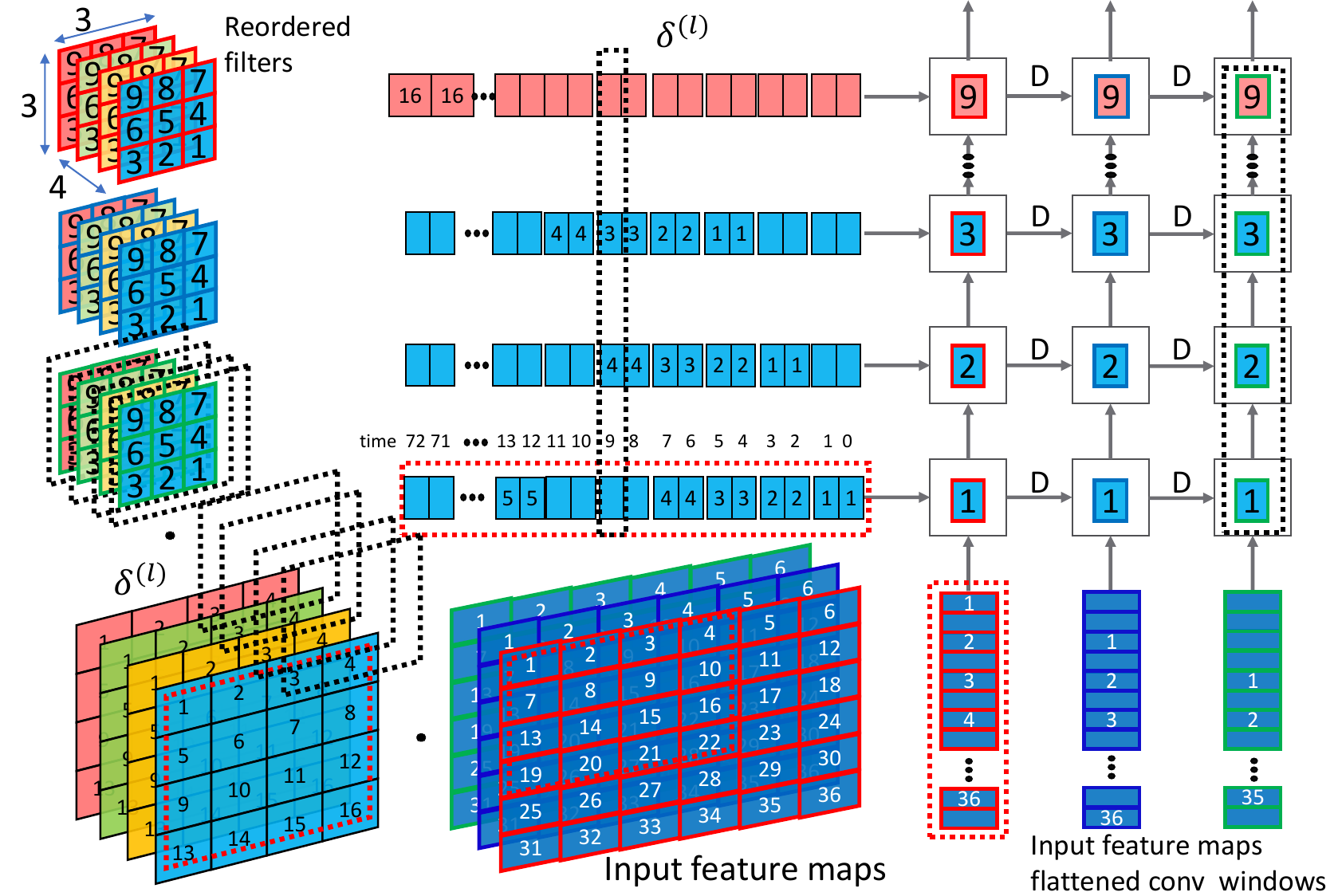}
	\vspace{-1pt}
	\caption{Interleaving of operations for the convolutional layer. The convolution windows for $G$ and $\delta$ calculations line up when mapped to OS and WS modes, respectively. The black dotted lines and red dotted lines represent the convolution windows for WS and OS, respectively. Taken from \cite{intergrad}}.
    \label{keshab-fig:sys_convp}
	\vspace{-0.25in}
\end{figure}

Gradient interleaving is developed in ~\cite{intergrad} based on \textit{interleaved scheduling} \cite{interleaving} to maximize the utilization of variables loaded into the array and enable a reduction in the number of on-chip memory accesses.
There are three avenues of reuse that can be exploited. Reuse of $\delta^{(l)}$ between computing $\boldsymbol{\delta}^{(l-1)}$ and $\boldsymbol{W}^{(l)}$, reuse of $W^{(l)}$ when computing $\boldsymbol{\delta}^{(l-1)}$ and updating the weights, and reuse of the result $G^{(l)}$  when updating the weights.
To enable these extra dimensions of reuse, {\em configurable systolic arrays} that can perform the matrix-matrix or matrix-vector operations in multiple configurations are necessary. 
Prior approaches have attempted to exploit this reuse at the cache level; however, gradient interleaving is the first attempt to systematically exploit this reuse at the array level. This approach outperforms the best traditional dataflow schemes by a factor
of $1.4\times \sim 2.2\times$ in terms of cycles and by a factor of up to $1.9\times$ in terms of memory accesses in the fully-connected layers.

Fig.~\ref{keshab-fig:sys_convp} highlights one possible interleaving method for the backward pass operations and how they can be mapped to a single systolic array.
The calculation of $\delta^{(l-1)}$ is performed in a weight-stationary manner (as highlighted by the black dashed line box) where the filter weights are stored in the systolic array.
$\delta^{(l)}$ is input from the array's left edge. This is a weight-stationary mapping with padded $0s$ to match the padding of the operation. 
A close observation of the dataflow pattern shows that each row of input $\delta^{(l)}$ is in the same format as required by the convolution window for calculating $G^{(l)}$. The critical observation is that the padded $0s$ introduced at the boundaries align with the part of $a^{(l-1)}$ not in the convolution window. Thus, this artificially creates the convolution windows (highlighted by the red dashed boxes) required for calculating $G^{(l)}$ in an output-stationary manner. Along the bottom edge, the previous layer's input, or input feature map, $a^{(l-1)}$, is unrolled and broadcast to the column. 
The interleaving of these operations can be easily achieved by holding $\delta^{(l)}$ and $a^{(l-1)}$ for two consecutive cycles. In addition, the gradient is calculated in the same PE where its corresponding weight is stored, thus allowing for in-place weight update.

Further, pipeline parallelism, like PipeDream~\cite{Pipedream} from Microsoft or GPipe~\cite{Gpipe} from Google, can be used to pipeline the backpropagation algorithm and to maintain high throughput while reducing model size at each stage.
The development of pipeline parallelism methods is analogous to the development of the original delayed least mean square algorithm \cite{DLMS}.
The bottleneck with current pipelining techniques is that they are limited to layer-level granularities or break up operation dimensions. Operation-level core designs allow fine-grained pipelining to balance workloads across multiple cores more effectively. Prior works on LayerPipe~\cite{layerpipe} and LayerPipe2 \cite{LayerPipe2} show that operation-level pipelining and retiming
enable fine-grained pipelines that better balance neural network workloads.
Prior evaluations compare scheduling algorithms for balanced workloads, including LayerPipe and PipeDream, on VGG16 and ResNet50. The tests were performed by sweeping the systolic array size from $32\times32$ to $256\times256$, and a minibatch size from $16$ to $256$ in powers of 2.
LayerPipe achieves on average $43\%$ improvement over PipeDream. In addition, it achieves greater than $80\%$ improvement with nine processors. LayerPipe consistently outperforms PipeDream across all systolic array sizes and batch sizes with only a minimal communication overhead.

\begin{figure}
	\centering
	\includegraphics[scale=0.3]{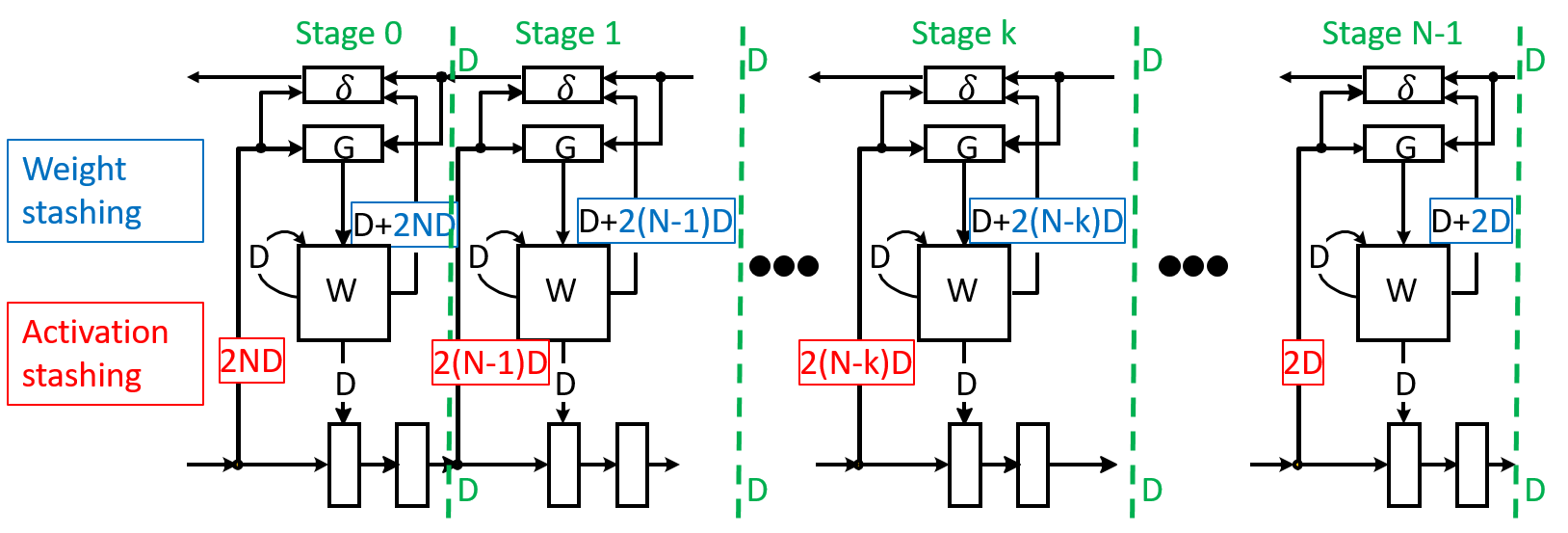}
	\vspace{-1pt}
	\caption{The growth of weight and activation stashing with number of pipeline stages \cite{layerpipe}.}
    \label{keshab-fig:deep_pipe}
	\vspace{-0.25in}
\end{figure}

The traditional and proposed methods to solve the inter-layer pipelining bottleneck exploit the use of delayed or stale gradients. However, the use of these delayed gradients requires {\em stashing} or storing intermediate results for weights and activations. This can lead to a significant increase in memory size for very deep pipelines.
For example, for each pipeline stage after a layer, the number of memory locations required is twice the number of layers to the right, including the current layer, as shown in Fig.~\ref{keshab-fig:deep_pipe}. 
This can become untenable in larger pipelines and necessitates techniques to mitigate this increase. Mitigating the activation stashing has been explored in prior works~\cite{replay1,replay2} with the use of {\em activation replays}. Activation replays trade off memory for computation by only storing the activation values at the stage boundaries. The intermediate layer activation values are \textit{recalculated} in the backward pass when a stage contains multiple layers. 

\subsection{Quantization, Sparsity, and Low-Rank Approximation}

Quantization cuts energy by reducing word-lengths for weights, activations, and intermediate states so that memory traffic and multiply-accumulate (MAC) cost drop dramatically, but it works best when tailored to data distributions and hardware~\cite{yirancasm2025}. Practical strategies include post-training quantization with robust calibration (percentile clipping, entropy/KL divergence, mean squared error) and quantization-aware training to learn scale factors and resilience to rounding \cite{ashkboos2024efqat,chen2025efficientqat,wang2025optimizinglargelanguagemodel,lang2024comprehensive}. Mixed precision is key: sensitive layers (embeddings, first/last, normalization) require higher precision while most computations require 8  or 4 bits. We can leverage the power of two (log-scale) or block floating point to simplify scaling, and use optimal or learned rounding to minimize quantization error. For transformers, we can quantize attention projections aggressively and compress key-value caches with fewer bits. 

Prior work on PermDNN \cite{permdnn} has demonstrated why structured sparsity matters. Instead of unstructured zeros that require heavy indexing and irregular memory access, PermDNN composes layers from permuted diagonal matrices, yielding predictable dataflow, lower memory movement, and simpler arithmetic while retaining expressive capacity through learnable permutations. That same design principle translates naturally to Transformers: attention and multi-layer perceptron (MLP) weights benefit from block- or N:M-structured sparsity. In attention, structured sparsity complements locality- or block-sparse patterns for long sequences, while in MLPs it aligns with grouped generalized matrix multiplications and quantization. The trade-off is choosing patterns rich enough to maintain accuracy while regular enough to stay fast. 

Low-rank approximation compresses dense weight matrices by factoring them into products of thinner matrices, exploiting the observation that many learned parameter tensors have low intrinsic dimensionality \cite{hu2021lora,liang2025qrlora}. Prior work on tensor decomposition \cite{liuCASM2023} has demonstrated various approaches to reducing the size of the model. Benefits include lower memory and compute, faster I/O, and an implicit regularization that can improve generalization.

\begin{figure}
    \centering
    \includegraphics[width=0.6\linewidth]{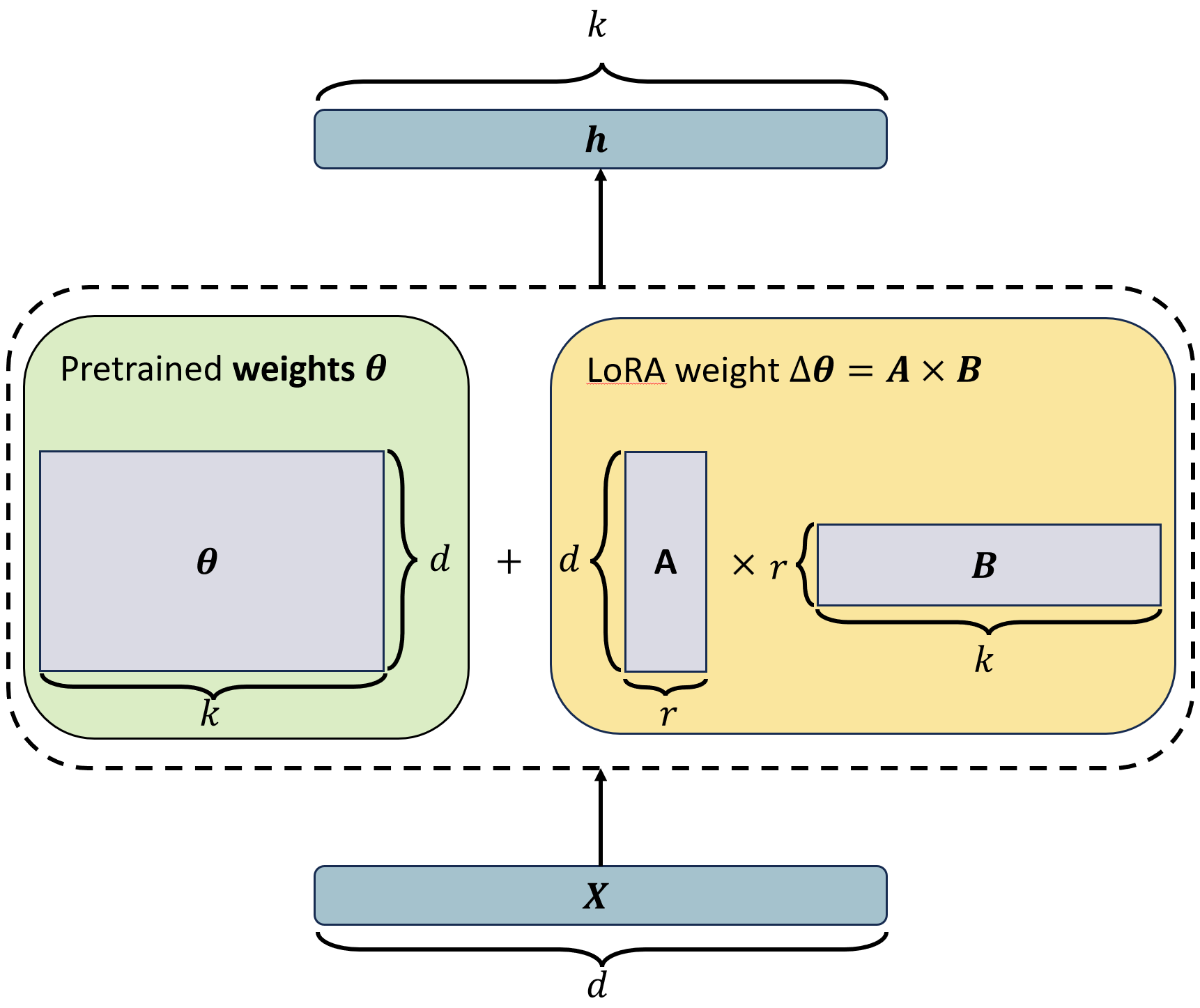}
    \caption{Low-Rank Adaptation (LoRA) for parameter-efficient fine-tuning. A pretrained weight matrix remains frozen while a low-rank update $\Delta\theta=AB$ is learned through two small trainable matrices.}
    \label{fig:LoRA}
\end{figure}

Among these low-rank methods, Low-Rank Adaptation (LoRA)~\cite{hu2021lora} has become one of the most practical techniques for parameter-efficient fine-tuning (PEFT) of LLMs. Instead of updating the full pretrained weight matrix, LoRA freezes the original model parameters and learns a pair of low-rank matrices that approximate the weight update as shown in Fig~\ref{fig:LoRA}. As a result, only a small fraction of the model parameters needs to be optimized, substantially reducing training memory, computation, and storage requirements while maintaining competitive task performance. Modularity is an important advantage of LoRA for domain-specific AI systems. Different LoRA modules can be trained to capture different functionalities within the same application domain. During deployment, these lightweight LoRA modules can be dynamically loaded, composed, or replaced while sharing a common foundation model. This modular adaptation reduces retraining costs, facilitates continual updates as domain knowledge evolves, and enables efficient specialization without maintaining multiple copies of pretrained LLMs.

\subsection{Mixture of Experts}

Mixture of Experts (MoE) plays a critical role in scaling AI models efficiently by introducing conditional computation. Instead of activating all parameters for every input, MoE architectures maintain a large set of specialized “expert” subnetworks and use a gating mechanism to select only a small subset (e.g., 2–4 experts) per token or sample. This approach allows models to dramatically increase parameter count without proportional increases in compute cost, enabling higher capacity for representation and generalization \cite{nikolic2025exploring}. Recent innovations like Switch Transformers and GLaM demonstrate that MoE can achieve state-of-the-art performance with lower inference cost, making it a cornerstone for building scalable and efficient AI systems \cite{du2022glam,fedus2022switchtransformer}.

\subsection{Model–Architecture Codesign}

The five research directions described in this section are not independent optimization techniques, but complementary components that collectively enable model--architecture co-design for compute-efficient AI systems. Rather than relying on a single optimization strategy, model--architecture co-design jointly considers algorithmic innovations and hardware implementations to maximize overall computational efficiency.

As shown in Fig.~\ref{fig:codesign}, each direction contributes to different aspects of the model-architecture codesign. HDC and VSA provide novel representations that reduce computation and memory movement. Reinforcement learning and knowledge distillation enable compact models to acquire strong reasoning capabilities while adaptively allocating computation. Energy-efficient training reduces model optimization cost through improved scheduling, data reuse, and parallel execution. Quantization, sparsity, and low-rank adaptation reduce memory traffic and arithmetic complexity, while mixture-of-experts (MoE) introduces conditional computation by only activating the model components required for each input. These techniques are working jointly for model--architecture codesign across the entire computing stack, including algorithms, model architectures, compiler optimizations, and hardware implementations, which provides a practical path toward energy-efficient domain-specific AI systems.

\section{Emerging AI Architectures for Memory-Augmented Domain-Specific Systems}\label{sec: Architecture}
The prior section has introduced computation mechanisms that can reduce energy consumption. In this section, we address three bottlenecks that determine whether small domain-specific models can compete with frontier general models: long-context processing without quadratic attention, structured memory without storing all knowledge in model parameters, and agent loops for decision-making. It is well known that the transformer complexity with respect to time and memory grows quadratically and linearly, respectively, with token length. This section describes new and emerging AI models that overcome complexity and memory bottlenecks of the transformer. We also describe emerging models for multimodal processing based on cognitive architectures. 

\subsection{Long-Context Efficiency: Sublinear Attention and State-Space Models}
Linear attention and state-space models enable linear-time, constant-memory sequence modeling. In recent work on log-linear attention, the fixed-size hidden state is replaced with a logarithmically growing set of hidden states \cite{guo2025loglinearattention}. State-space models like Mamba can overcome the limitations of transformers in handling very long sequences \cite{gu2024mamba,dao2024mamba2}. Unlike transformers, which rely on quadratic-cost attention mechanisms, Mamba uses structured state-space dynamics to model sequences in linear time, making it highly efficient for tasks like language modeling, audio processing, and genomics. At its core, Mamba replaces attention with a selective state-space mechanism that dynamically updates a hidden state using equations inspired by control theory. This approach eliminates the need for large key-value caches. Mamba introduces innovations such as time-varying parameterization, hardware-aware parallel scan algorithms, and kernel fusion, enabling up to 5× faster inference and linear scaling with sequence length compared to transformers. Previously proposed efficient attention mechanisms are summarized in Table~\ref{tab:table 1}.

\begin{table*}
\centering
    \caption{Summary of previously proposed efficient attention mechanisms under the unified formulation~\cite{guo2025loglinearattention}: $\mathbf{O} = (\mathbf{A} \odot \mathbf{M}) \mathbf{V}$, where $\mathbf{M}$ is a lower-triangular (causal) matrix. For notational brevity, the symbol $\mathcal{T}_K(\mathbf{A}) = (\mathbf{A} \odot \mathbf{L})(\mathbf{I}+\mathbf{KK}^T\odot(\mathbf{I} - \mathbf{L}))^{-1}$ with lower-triangular matrix of 1s $\mathbf{L}$. The decoding time is the time per step, and decoding space refers to the overall memory complexity during generation. Here, $\mathbf{K,Q,V} \in R^{T\times d}$ refer to the key, query, and value matrices for a given input token sequence of length $T$, and embedding dimension $d$.}
    \begin{tabular}{c|c c c c c}
    \hline
        Model & \textbf{A} & \textbf{M} (Data Dependent?) & Training Algorithm/ Time & Decoding Time & Decoding Space \\
        \hline
        Attention & $\sigma(\mathbf{QK}^T)$ & Mask (\textbf{N}) & FlashAttention $O(T^2)$ & $O(T)$ & $O(T)$\\
        \hline
        Linear Attention~\cite{katharopoulos2020transformersrnnsfastautoregressive} & $\mathbf{QK}^T$ & Mask (\textbf{N}) & Chunk-recurrent $O(T)$ & $O(1)$ & $O(1)$\\
        \hline
        RetNet~\cite{sun2023retentivenetworksuccessortransformer} & $\mathbf{QK}^T$ & Semiseparable (\textbf{N}) & Chunk-recurrent $O(T)$ & $O(1)$ & $O(1)$\\
        \hline
        Mamba-2~\cite{dao2024mamba2} & $\mathbf{QK}^T$ & Semiseparable (\textbf{Y}) & Chunk-recurrent $O(T)$ & $O(1)$ & $O(1)$\\
        \hline
        Multi-Hyena~\cite{massaroli2023laughinghyenadistilleryextracting} & $\mathbf{QK}^T$ & Toeplitz (\textbf{N}) & FFT $O(T\log T)$ & $O(\log^2 T)$ & $O(T)$\\
        \hline
        DeltaNet~\cite{schlag2021lineartransformerssecretlyfast,yang2025parallelizinglineartransformersdelta} & $\mathcal{T}_K(\mathbf{QK}^T)$ & Mask (\textbf{N}) & Chunk-recurrent $O(T)$ & $O(1)$ & $O(1)$\\
        \hline
        Gated DeltaNet~\cite{yang2025gateddeltanetworksimproving} & $\mathcal{T}_K(\mathbf{QK}^T)$ & Semiseparable (\textbf{Y}) & Chunk-recurrent $O(T)$ & $O(1)$ & $O(1)$\\
        \hline
        Log-linear Mamba-2~\cite{guo2025loglinearattention} & $\mathbf{QK}^T$ & Hierarchical (\textbf{Y}) & Chunk-Scan $O(T\log T)$ & $O(\log T)$ & $O(\log T)$\\
        \hline
        Log-linear Gated DeltaNet~\cite{guo2025loglinearattention} & $\mathcal{T}_K(\mathbf{QK}^T)$ & Hierarchical (\textbf{Y}) & Chunk-Attention $O(T\log T)$ & $O(\log T)$ & $O(\log T)$\\
        \hline
    \end{tabular}
    \label{tab:table 1}
\end{table*}

\subsection{Structured Domain Knowledge: Knowledge Graphs}
A knowledge graph (KG) provides compositional structure, where domain primitives are represented as head-relation-tail edges, and their paths encode higher-level concepts. Recent work has proposed a task generation pipeline to synthesize tasks directly from KG primitives, enabling models to acquire and compose them for reasoning \cite{dedhia2025}. Language models can then be fine-tuned on the resultant KG-grounded curriculum to demonstrate domain-specific superintelligence. While broadly applicable, this approach has been validated in medicine, where reliable KGs exist. Using a medical KG, 24,000 reasoning tasks paired with thinking traces derived from diverse medical primitives are curated, and the QwQ-32B model is fine-tuned on this curriculum to obtain QwQ-Med-3 as a step towards medical superintelligence. This finding indicates the future that broad AGI-like behavior may emerge through the composable interaction of efficient domain-specific agents, rather than through a single monolithic model.

\subsection{Modality-General Processing: Perceiver IO}
Present neural network models are tailored to a single specific task, and they demonstrate impressive performance in that particular task. However, real-world applications require a framework that generalizes across different input modalities and output tasks. Conventional multimodal NNs\cite{NIPS_alayrac,kaiser2017modellearn,lu_pretrained_XFMR} process inputs of different modalities independently with domain-specific architectures and integrate them together with a separate fusion NN. This approach, nevertheless, is inefficient as model complexity increases with the number of inputs, outputs, and attention span of traditional transformer architecture is strictly restricted to its context window (of the order of thousands). Inspired by the previous work on \textit{Perceiver} \cite{perceiver}, the authors proposed \textit{Perceiver IO}\cite{perceiverIO}, a unified processing framework for arbitrary input modalities and functionality. \textit{Perceiver IO} unravels the scalability issue of attention mechanism in vanilla transformer\cite{vaswani2017attention} architecture by performing multimodal processing in a fixed-dimensional latent space, and this generalized pipeline doesn’t require any domain-specific data manipulation. Moreover, the approach has been tested on a variety of tasks and modalities of input, ranging from natural language processing without tokenization, audio-visual sequence processing, optical flow estimation without specific architectural features, image classification, or on symbolic unordered sets. On top of that, perceiver IO demonstrates impressive ability to attend non-homogeneously with an increased attention context window of the order of hundreds of thousands, without increasing the computational or memory complexity. 

In essence, the Perceiver IO architecture can be characterized by three fundamental components: encode, process, and decode. Inputs are assumed to be a simple two-dimensional byte array, a set of elements that consists of pixels, patches of images, characters, words, or learned embeddings. Input arrays are encoded into a latent space by the attention module, and processing is performed in the latent vector space by a sequence of attention modules. Output vectors are generated by decoding from the latent space with the help of the output query vector, which dictates the model to produce output for specific tasks. Unlike the transformer attention mechanism, Perceiver IO's computational complexity is not quadratically dependent on input or output dimensions due to processing in the fixed latent vector space, and allows the consolidation of arbitrary modalities in the same model architecture. 

\subsection{Test-Time Memory: Titans}
Traditional transformer\cite{vaswani2017attention} architecture memory is limited in terms of context length, and the key value associative memory embedded into the attention mechanism is susceptible to poor generalization and length extrapolation, and suffers in long spans of reasoning tasks\cite{length_extrapolation,length_extrapolation_2}. However, the context length of transformer architecture is not scalable due to quadratic time/memory complexity and transformer memory is restricted to the tokens in the context window - no presence of persistent memory for the older sequences. This is dissimilar to the way human memory and cognitive system functions, where short-term memory, long-term memory, and input-agnostic meta-memory operate independently in current context. Titans\cite{titans} proposes a new memory architecture that utilizes a novel neural long-term memory (LTM) along with attention mechanism acting as the short-term memory and a simple MLP acting as the input-agnostic task knowledge memory. Subsequently, next question that arises is how to incorporate these independent memory segments into a functional memory block and the authors of this work presented three approaches in this situation - memory as a context, memory as a layer, and memory as gating. Experiments performed on various benchmarks on natural language processing, reasoning, genomics, time series, and needle-in-haystack tasks demonstrate superior performance along with an effective context window in the order of millions of tokens. 

The main purpose of neural LTM is to encode abstractions from the historical data. Although transformer-based LLM’s tend to excel in memorizing training data\cite{memorize_training_data1,memorize_training_data2,memorize_training_data3}, this phenomenon interferes with the models’ generalization capabilities and results in poor performance at test time. In order to train the long-term neural memory in test time, training is designed to be an online learning with a memory meta-model. Similar to human memory, if an event is beyond expectations or surprising, that becomes more memorable and this effect has been emulated in the NN by considering the gradient as surprise signal to evaluate the novelty of the information. This surprise-based updating is performed at the test time to avoid overfitting to the training data, and the memory module is updated according to the following equation 
\begin{equation}
    \mathcal{M}_{t} = \mathcal{M}_{t-1} - \theta_{t} \nabla l(\mathcal{M}_{t-1}; x_{t})
\end{equation}

Here, $\mathcal{M}_{t}$ is the neural LTM module, $x_t$ is the input at time step $t$, and $\nabla l(\mathcal{M}_{t-1}; x_{t})$ is the surprise metric. However, updating memory solely based on the surprise score can result in disregarding follow-up details after a surprising event and this can be solved by segmenting the surprise metric into past surprise and momentary surprise, similar to introducing momentum to the surprise metric across time steps.
\begin{equation}
    \mathcal{M}_{t} = \mathcal{M}_{t-1} + \eta_{t}\cdot \mathcal{S}_{t-1} - \theta_{t} \nabla l(\mathcal{M}_{t-1}; x_{t})
\end{equation}
Note that, $\eta_{t}$ is a data-dependent surprise decay factor(dependent on $x_t$), $\mathcal{S}_{t-1}$ as the past surprise factor, $\nabla l(\mathcal{M}_{t-1}; x_{t})$ represents the momentary surprise factor, and $\theta_t$ is the weight factor controlling the contribution of momentary surprise in updating the LTM.
Also, keeping in mind the risk of memory overflow while dealing with very large sequences, a forgetting mechanism has been developed to discard unnecessary elements of memory. On the other hand, persistent memory or meta memory has been modeled as a set of learnable but input-independent parameters pre-appended to the input token sequence. From the perspective of the memory architecture, this set of parameters stores the input-agnostic abstraction of task knowledge while neural LTM acts as the contextual memory, which depends on the input sequences. Concatenation of persistent memory parameters at the front end of the input sequence also improves the performance of the attention mechanism by distributing attention weights along the context window. Both neural long-term memory and persistent memory can be implemented using an MLP, but weights in the persistent memory module are made input-independent.

\subsection{Agent Loops: Cognitive Architectures for Language Agents}

\begin{figure}
    \centering
    \includegraphics[width=\linewidth]{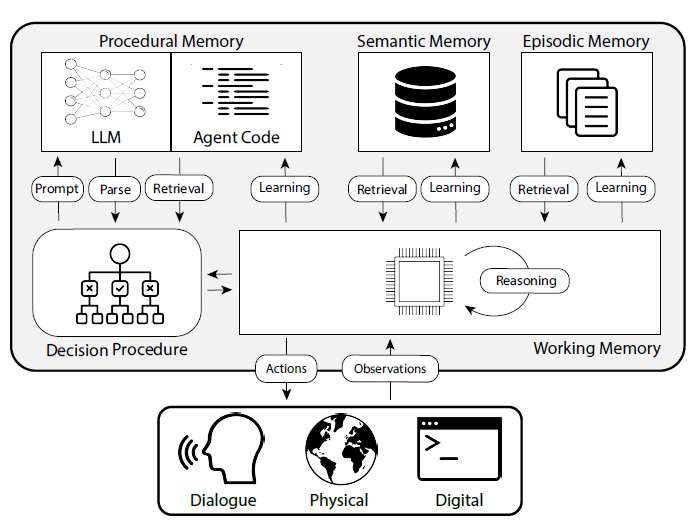}
    \caption{Cognitive Architecture for Language Agents (CoALA). The framework unifies LLM-based agents through modular memory, decision-making, and action components. The architecture supports iterative reasoning, retrieval, learning, and interaction with both internal memory and the external environment.~\cite{coala}}
    \label{fig:CoALA}
\end{figure}

Many AI systems based on large language models use those models as a foundation for creating language agents, with components that interact with one another and the world. A generalized framework is needed for describing such language agents so that different agent structures can be compared, limitations of current research can be identified, terminology can be clarified, and future agents can be built from modular components. Drawing inspiration from the cognitive architectures\cite{acquiring_grounding,soar,soar_introduction,mapping_AGI}, where agents are dictated by hand-crafted productions or set of rules, the
Cognitive Architecture for Language Agents (CoALA)~\cite{coala} approach treats language models as probabilistic production systems. Similar to traditional production systems, which map one symbolic state to another, LLMs produce a string given an input prompt with a learned probability distribution. As shown in Fig.~\ref{fig:CoALA}, CoALA~\cite{coala} proposes a unified framework to characterize and generalize language agents with three core dimensions: a modular memory component, an action space to interact with internal memory and external environment, and a looped decision-making process involving observation, planning, execution, and learning stages. In summary, the CoALA~\cite{coala} framework acts as a foundational roadmap to consolidate traditional structured, rule-based cognitive architectures and modern probabilistic behavior of LLMs. 

The memory component of the CoALA~\cite{coala} framework consists of short-term working memory and LTM with three elements: episodic memory, semantic memory, and procedural memory. Short-term working memory contains the agent’s current observation, goals, and intermediate reasoning results, while episodic LTM encodes agent’s past experience. Semantic memory consists of facts or world knowledge and procedural memory stores common task knowledge encompassing implicit skills encoded in the weights of the language model or decision-making capability of the model. The second component is the action space and it is defined as the set of possible actions the agent can take, and the authors of CoALA~\cite{coala} framework categorize the action space of an agent into external grounding actions, where the agent interacts with the external environment by transforming environmental feedback into texts and internal actions altering the internal state and memories. Within the internal action space, the agent can retrieve information from LTM into working memory, learn new information, and update the episodic, semantic, or procedural memory, or perform reasoning over working memory. The process of making decisions in the CoALA~\cite{coala} framework is conceived as a repeated cycle, starting with an observation encoded into working memory. Subsequently, the agent goes through a planning stage where the agent utilizes retrieval of LTM and reasoning to propose actions and evaluate the possible actions through heuristics, value functions or further LLM reasoning. Finally, the agent executes the action and updates the internal memory as well as receives feedback from the environment and the cycle continues. This modular approach is employed in diverse levels of applications by varying the memory, action space, and decision-making procedures, ranging from robotics, reasoning, social simulacra, and ablation study on Saycan\cite{saycan1}, ReAct\cite{react_paper}, Voyager\cite{voyager_paper}, Tree of thoughts\cite{yao2024tree}, and so forth. 

\section{Conclusions}

This paper provides a vision that many critical applications can be served more effectively by energy-efficient domain-specific AI systems than by continually upscaling general-purpose LLMs. The proposed roadmap integrates compact models with reasoning mechanisms, multimodal grounding, external knowledge, efficient training, model compression, conditional computation, and specialized hardware. In this architecture, capability comes from the composition of smaller parametric models with trusted knowledge and explicit reasoning, rather than from storing all knowledge and skills in giant models. The expected $\geq{1000\times}$ energy improvement should be interpreted as a system-level target, not as a guaranteed result of any single method. It becomes plausible through cumulative gains from smaller models, fewer training tokens, retrieval and structured knowledge, low-precision and sparse computation, reduced memory movement, and workload-specific hardware. The key challenge is to empirically validate this path while preserving reliability, verifiability, and domain coverage. If validated, this path could provide a foundation for future brain-like intelligence that is both capable and energy efficient.

\section{Acknowledgment}
The authors thank Mohamed Mejri of Georgia Institute of Technology for his help in preparation of this paper.

\newpage
\bibliographystyle{IEEEtran}
\bibliography{reference}
\end{document}